\documentclass[preprint,12pt]{elsarticle}

\usepackage{amssymb}
\usepackage{amsmath}
\usepackage{amsthm}

\newtheorem{Theorem}{Theorem}
\newtheorem{Lemma}{Lemma}
\newtheorem{Definition}{Definition}

\journal{Photography, Image Processing, and Computer Vision }

\begin{document}
	
	\begin{frontmatter}
		
		
		
		\title{Modeling Image Tone Dichotomy with the Power Function}
		
		
		\author{Axel Martinez} 
		\author{Gustavo Olague\corref{cor1}}
		\author{Emilio Hernandez}
		\cortext[cor1]{corresponding author Gustavo Olague, email: olague@cicese.mx}
		\affiliation[label1]{organization={CICESE, Computer Science},
			addressline={Carretera Tijuana-Ensenada 3918, Zona Playitas}, 
			city={Ensenada},
			postcode={22760}, 
			state={Baja California},
			country={Mexico}}
		
		\begin{abstract}
			The primary purpose of this paper is to present the concept of dichotomy in image illumination modeling based on the power function. In particular, we review several mathematical properties of the power function to identify the limitations and propose a new mathematical model capable of abstracting illumination dichotomy. The simplicity of the equation opens new avenues for classical and modern image analysis and processing. The article provides practical and illustrative image examples to explain how the new model manages dichotomy in image perception. The article shows dichotomy image space as a viable way to extract rich information from images despite poor contrast linked to tone, lightness, and color perception. Moreover, a comparison with state-of-the-art methods in image enhancement provides evidence of the method's value.
		\end{abstract}
		
		\begin{graphicalabstract}
			\includegraphics[width=13.0cm]{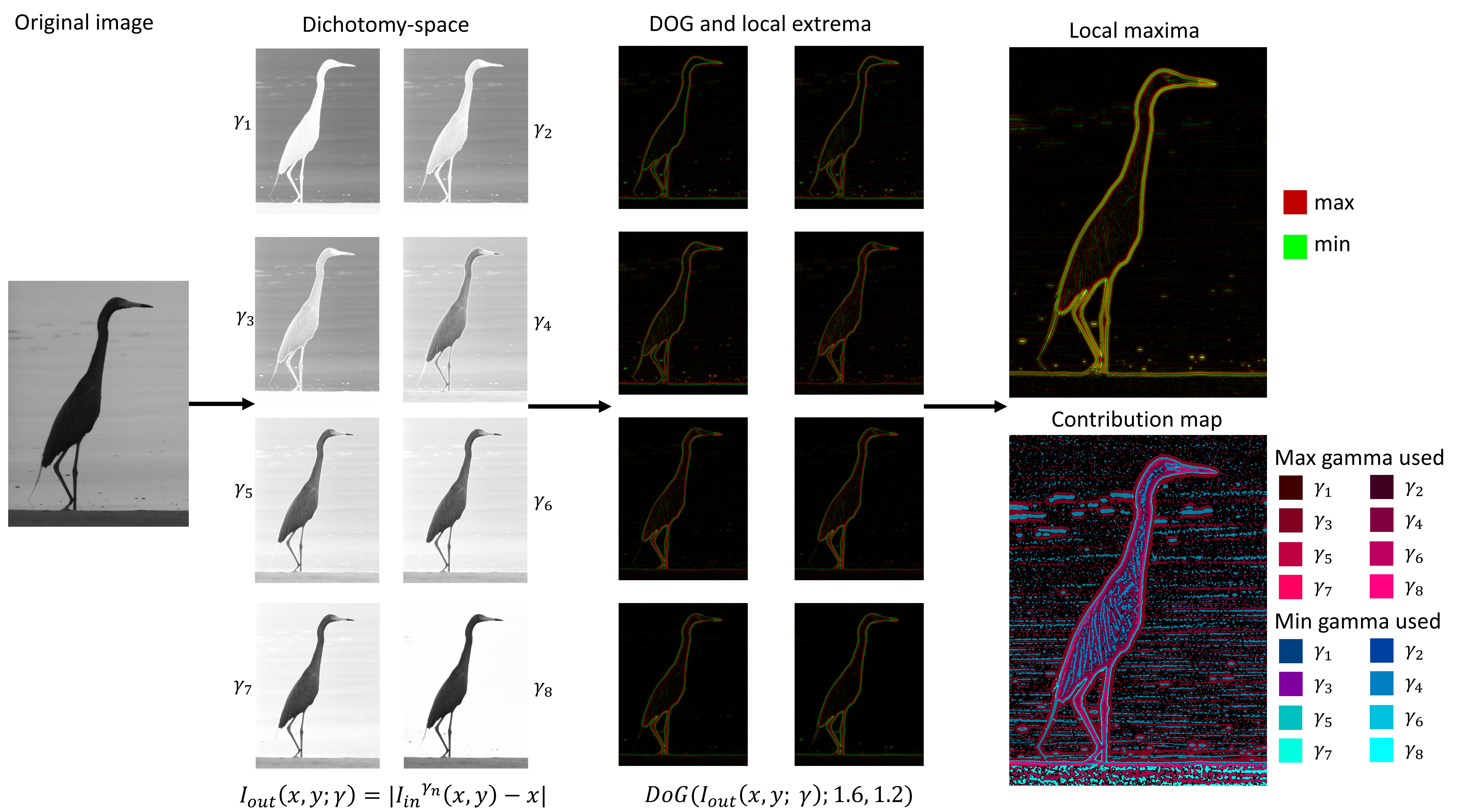}
		\end{graphicalabstract}
		
		\begin{highlights}
			\item This article introduces the mathematical concept of tone dichotomy.
			\item Mathematical results provide significant tools with practical implications for image processing and analysis.
			\item The article extends image tone dichotomy to multiple images, creating dichotomy space.
			\item Examples show the results on underexposed, overexposed, and a combination of both cases.
			\item Comparison with image enhancement methods provides evidence of the method's value against state-of-the-art in a particular benchmark.
		\end{highlights}
		
		\begin{keyword}
			
			
			
			Brightness \sep Lightness \sep Luminance \sep Image Contrast \sep Color Space \sep Dichotomy Space
			
		\end{keyword}
		
	\end{frontmatter}
	
		
		
\section{Introduction}
		
Photography means "drawing with light," derived from the Greek photo, which means light, and graph, which means to draw. Light is the most critical variable when conceiving a mathematical model, but it is still daunting how difficult it is to encapsulate their meaning. Light permeates analog and digital photography through the concepts of brightness, lightness, and luminance, affecting tone and other attributes of color appearance.
Classical approaches to regulate the amount of light in an image are made through gamma correction or gamma, a nonlinear operation used to encode or decode luminance, and defined by the following power law expression:
		
\begin{displaymath}
	V_{out} = AV_{in}^\gamma\;\;\;,
\end{displaymath}
		
\noindent
where the non-negative real input value $V_{in}$ is raised to the power $\gamma$ and multiplied by a constant $A$ to get the output $V{out}$. Usually, $A=1$ and domain and codomain are in the range $\{0\ldots1\}$. 
		
There are two power-law effects on an image: 1) encoding gamma if $\gamma < 1$ with a compressive nonlinear result known as gamma compression, and 2) decoding gamma if $\gamma > 1$ whose expansive nonlinear result derives the name gamma expansion. Both effects intend to improve human perception of brightness or lightness under common illumination conditions. This empirical relationship and its theory is known as Steven's power law, named after psychophysicist Stanley Smith Stevens \cite{ref-journal1} and has a close relationship with Weber-Fechner logarithmic function \cite{ref-journal2}. The aim is to make apparent significant individual differences in data. We keep on with the power function, although anyone with mathematical knowledge could easily derive the results exposed in this article for other functions. The goal in this paper is to present a new mathematical modeling based on the power function, which extends the capacities to represent image content from the viewpoint of tone perception and invariant information extraction. Application of gamma correction have been instrumental in many technologies such as: CRT (cathode ray tubes) of television, modern display devices, and computer graphics \cite{ref-book1}. In fact, the application in film photography known as sensitometry has its origin in the work of Ferdinand Hurter and Vero Charles Driffield (circa 1976) and the density-exposure curve, characteristic curve or H\&D curve \cite{ref-book1a}. 
		
The relevance of the power law for lightness modeling attracts the interest of scientists and engineers. For example, Gopalan and Arathy describe a fuzzy approach for image enhancement considering underexposed and overexposed regions \cite{ref-journal3}. The aim was to manipulate image properties like intensity, contrast, and number of gray levels through fuzzy sets. Yelmanov and Romanyshyn also work on the image enhancement problem \cite{ref-journal4}. This work applies an adaptive power-law image intensity transformation for real-time video sequences. Huang et al. introduced efficient contrast enhancement methods based on adaptive gamma correction with weighting distribution \cite{ref-journal5}. The method modifies histograms to improve the brightness of dimmed images.  In the work of Rahman et al., the goal is also image enhancement while attempting to manipulate image information through an adaptive gamma correction \cite{ref-journal6}. The authors highlight the limitations of image-capturing devices, producing that image quality may be degraded, and multiple factors that compromise the user's expectation of clear and soothing views. Hence, based on adaptive gamma correction, the proposed algorithm improves image contrast and produces natural illumination. Wang et al. apply gamma correction in combination with deep neural networks for low-light image enhancement \cite{ref-journal7}. Authors argue that exponential functions are computationally demanding and propose a simplification through Taylor expansion when integrating the operation within the network layers. As we can appreciate, all reviewed works on image enhancement through the power law attempt to obtain visually pleasing images, mainly when there is low-light image exposure. Such methods look for an optimal parameter without considering that extracting visual information requires multiple values, each improving a particular case: underexposed, overexposed, or a combination of both, regardless of image semantics. In another work, Moroz et al. present power function algorithms implemented in hardware \cite{ref-journal8}. That work highlights the critical importance of efficient hardware implementation of power functions for tasks such as accelerators of neural networks, power electronics, machine learning, computer vision, and intelligent robotic systems.
		
In a different research trend, scientists highlight the idea of dichotomy in image analysis. For example, Blakeslee and McCourt \cite{ref-journal9} criticize the theoretical approaches to lightness and perception presented by Gilchrist \cite{ref-journal10}. We observe that new mathematical reasoning about lightness, brightness, and contrast is necessary, and the present article aims to open new research avenues not only for the physics of image formation but also from the neuroscientific viewpoint. Indeed, there is only one work in computer vision where scientists use the term dichotomous for image segmentation based on empirical reasoning applying implicit modeling through a database \cite{ref-journal11}. There is no formal definition of the concept nor mathematical reasoning to describe dichotomy in image analysis in general. Other efforts to understand dichotomy from a mathematical perspective include the work of Brill and Carter, where they revisited the dichotomy between the logarithmic and power-law models of lightness perception  \cite{ref-journal12}. The work highlights the necessity of a new luminance scale-invariance to describe the equality of lightness differences correctly. Finally, Bi et al. proposed the existence of mechanisms in the natural visual system to comprehend central-peripheral dichotomy \cite{ref-journal13}. This article aims to present a mathematical analysis of tone image dichotomy that can be helpful not only in color but in time and space, as well as for practical photography and as a tool for studying natural and artificial systems. The main contributions of this article are:
		
\begin{enumerate}
\item The introduction of the concept of image tone dichotomy for computational photography.
\item The proposal of a mathematical model helpful to improve image information that is underexposed, overexposed, and a mixture of both cases where, in this last case, the dichotomy is more apparent.
\item The derivation of properties for image tone dichotomy following deductive reasoning with practical implications.
\item The reasoning for building image dichotomy space to extract illumination information is similar to scale space theory.
\item Illustrative examples to improve image information whose photographs were badly exposed independently of aperture, shutter speed, and ISO.
\item A comparison with state-of-the-art image enhancement methods shows the proposal obtains consistent best values in a standard test.
\end{enumerate}
		
Organization of the paper.
		
In Section \ref{Preliminaries}, we present the power function and the properties necessary to understand the mathematical reasoning. The main achievement of this work is in Section \ref{Dichotomy}. The concept of an image as the graph of a function \cite{ref-book2} serves to understand the analysis of tone dichotomy through gamma correction. The Theorem \ref{theorem} and Lemmas \ref{Lemma6} to \ref{Lemma14} provide the necessary tools to implement practical results. Section \ref{Results} offers three examples of underexposed, overexposed, and a mixture of both problems to illustrate the ability to extract visual information from images with poor contrast. We provide a complete test using two standard metrics on a low-light image enhancement problem to show that the proposed approach outperforms state-of-the-art methods. Finally, we close the paper with a discussion about the implications of this work in image formation, as well as a conclusion and future work.
		
\section{Preliminaries and Auxiliary Results}
\label{Preliminaries}
The power function is part of algebraic-real functions subdivided into rational and irrational functions \cite{ref-book3}. The power function basic equation is:

\begin{equation}
	f(x) = a x^n \;\;\; n = 1,2,3,\ldots
	\label{eq.1}
\end{equation}

\noindent
However, the most general algebraic power functions are written as follows:

\begin{equation}
	f(x) = x^{\frac{m}{n}} \;\;\; ,
\end{equation}

\noindent
note that such an equation could be interpreted as the n-th root of the m-th power:

\begin{equation}
	f(x) = x^{\frac{m}{n}} =  \sqrt[n]{x^m}\;\;\; .
	\label{eq.2}
\end{equation}

Remark 1. There are two main ways to understand the behavior of the power function\footnote{In signal processing, the power function is an element-wise transformation function where the values of $a$ and $n$ are always positive.}. According to Equation (\ref{eq.1}), two important cases are possible when $n$ is an integer: 1) if $n$ is even, the function behaves similarly to the quadratic function. 2) while if $n$ is odd, the function behaves similarly to the cubic function. Hence, the power functions of higher degrees describe curves that grow faster than quadratic or cubic functions. All functions at $x=0$ have a value of $0$, and at $x=1$, the value of $a$. Functions with odd $n$ come from negative infinite and at $x=0$ have n-fold zero, a saddle point, and later go into positive infinite.

Remark 2. According to Equation (\ref{eq.2}), we assume that $n > 0$ and that $m \neq 0$ take positive values ($m$ and $n$ are integers). If $m$ and $n$ are odd, the function is valid for all $x$, and it can have positive and negative values, but we restrict the domain to a small part of the first quadrant $0 \leq x \leq 1$. The case for $m$ and $n$ even does not apply since the fraction $\frac{m}{n}$ can still be reduced by two into an integer. If $m$ is even and $n$ is odd, the function's domain exists for all $x$ and does not have negative values. If $m$ is odd and $n$ is even, the function exists only for $x \geq 0$ and does not have negative values.

Properties of Equation (\ref{eq.1}) 

\begin{tabular}{lll}
	Periodicity: & & None\\
	Domain of definition: & &	$-\infty < x < \infty$ \\
	Range of values:	&	$n$ even & $a > 0 : \;\;\; 0 \leq f(x) < \infty$ \\
	& $n$ odd & $a \neq 0$ : \;\;\; $-\infty < f(x) <\infty$ \\
	Quadrant: & $n$ even & $a>0$  : \;\;\; first and second quadrants\\
	& $n$ odd& $a>0$: \;\;\;\; first and third quadrants\\ 
	Monotonicity: & $n$ even & $a>0$: \;\; $x\geq0$ strictly monotonically increasing\\ 
	& $n$ odd & $a>0$: \;\; strictly monotonically increasing\\
	Asymptotes: & $n$ even & $a>0$: \;\;\; $f(x) \rightarrow +\infty$ for $x \rightarrow \pm \infty$\\
	&$n$ odd & $a >0$: \;\;\; $f(x) \rightarrow \pm \infty$ for $x \rightarrow \pm \infty$\\
	Symmetries: & $n$ even & mirror symmetry about $x=0$\\
	& $n$ odd & point symmetry about $x=0, f(x)=0$\\
\end{tabular}

\noindent
Regarding particular values, note that all power functions with $n$ integer have a zero value of $x=0$. There are no discontinuities or poles. Also, there are no extreme values when $n$ is odd; for $n$ even, there is a minimum at $x = 0$ for $a>0$. Also, there is no inflection point for $n$ even, with a saddle point at $x=0$ when $n$ is odd.

Properties of Equation (\ref{eq.2}) 

\begin{tabular}{lll}
	Periodicity: & & None\\
	Domain of definition: &$n$ even &	$0 \leq x < \infty$ \\
	& $n$ odd & $-\infty < x < \infty$\\
	Range of values:	&	$m$ or $n$ even &  $0 < f(x) < \infty$ \\
	& $m$ or $n$ odd & $-\infty < f(x) <\infty$ \\
	Quadrant: & $m$ even & first and second quadrants\\
	& $n$ even & first quadrant\\
	& $m$ and $n$ odd &  first and third quadrants\\ 
	Monotonicity: & $x>0$ and $m>0$:  & strictly monotonically increasing \\ 
	& $x<0$: & according to the symmetries \\
	Symmetries:& $m$ even & mirror symmetry about the y-axis\\
	& $n$ even & no symmetry\\
	& $m$ and $n$ odd & point symmetry about the origin\\
	Asymptotes: & $m>0$: &$f(x) \rightarrow \infty$ as $x \rightarrow \infty$\\
\end{tabular}

\noindent
Remark 3. The particular values of Equation (\ref{eq.2}) are as follows: in the case of $m>0$, the function has a zero at $x=0$, the function has no discontinuities, and it has no poles when $m>0$. Regarding extrema, there is a minimum when $x=0$ for $m$ even and $m>n$. Finally, in the case of points of inflection, there is a saddle point at $x=0$ when $m$ and $n$ are odd, and $m>2n$.

\begin{itemize}
	\item Reciprocal of function. 
	
	The reciprocal function of Equation (\ref{eq.1}) is a hyperbola of power $n$:
	
	\begin{displaymath}
		\frac{1}{f(x)} = \frac{1}{ax^n} = \frac{1}{a}x^{-n} \;\;\;,
	\end{displaymath}
	
	\noindent
	while for Equation (\ref{eq.2})
	
	\begin{displaymath}
		\frac{1}{f(x)} = \frac{1}{x^{\frac{m}{n}}} = x^{\frac{-m}{n}} \;\;\;.
	\end{displaymath}
	
	\item Inverse function
	
	The inverse function of Equation (\ref{eq.1}) is:
	\begin{displaymath}
		f^{-1}(x) = \sqrt[n]{\frac{x}{a}} \;\;\;.
	\end{displaymath}
	
	\noindent
	Note that if $n$ is odd the function can be inverted for $-\infty < x < \infty$, while if $n$ is even, the function can only be inverted for $x \geq 0$.
	
	In the case of Equation (\ref{eq.2}) the inverse function has a reciprocal value in the exponent:
	
	\begin{displaymath}
		f^{-1}(x) = x^{\frac{n}{m}} \;\;\;.
	\end{displaymath}
	
	\noindent
	This property is important in case we wish to revert the image transformation.\\
	
	\item Conversion formulas of the power function. 
	
	There are some conversion formulas of Equation (\ref{eq.1}).
	
	Recursion relation
	
	\begin{displaymath}
		ax^n = x \cdot ax^{n-1} \;\;\;.
	\end{displaymath}
	
	The sum of two power functions:
	
	\begin{displaymath}
		ax^n \pm bx^n = (a\pm b)x^n \;\;\;.
	\end{displaymath}
	
	The product of a power function of degree $n$ and a power function of degree $m$ is a power function of degree $n+m$.
	
	\begin{displaymath}
		ax^n \cdot bx^m = (a\cdot b) x^{n+m} \;\;\;.
	\end{displaymath}
	
	The quotient of a power function of degree $n$ and a power function of degree $m$ is a power function of degree $n-m$.
	
	\begin{displaymath}
		\frac{ax^n}{bx^m} = \frac{a}{b} x^{n-m} \;\;\;.
	\end{displaymath}
	
	The m-th power of a power function of degree $n$ is a power function of degree $n\cdot m$.
	
	\begin{displaymath}
		(ax^n)^m = a^m x^{n\cdot m} \;\;\;.
	\end{displaymath}
	
	Some conversion formulas for Equation (\ref{eq.2}) are:
	
	Product of two functions:
	
	\begin{displaymath}
		x^a \cdot x^b = x^{a+b} \;\;\;,\;\;\; a=\frac{m_1}{n_1}, b=\frac{m_2}{n_2} \;\;\;.
	\end{displaymath}
	
	Quotient of two functions:
	
	\begin{displaymath}
		\frac{x^a}{x^b} = x^{a-b} \;\;\;,\;\;\; a=\frac{m_1}{n_1}, b=\frac{m_2}{n_2} \;\;\;.
	\end{displaymath}
	
	Power of a function:
	
	\begin{displaymath}
		(x^a)^b = x^{a\cdot b} \;\;\;,\;\;\; a=\frac{m_1}{n_1}, b=\frac{m_2}{n_2} \;\;\;.
	\end{displaymath}
	
	\item Derivative of power law. 
	
	The derivative of Equation (\ref{eq.1})
	
	The derivative of a power function of degree $n$ is a power function of degree $n-1$:
	
	\begin{displaymath}
		\frac{d}{d x} a x^n = n ax^{n-1} \;\;\;,
	\end{displaymath}
	
	\noindent
	while the k-th derivative of  power function of degree $n$ is:
	
	\begin{displaymath}
		\frac{d^k}{dx^k} ax^n = \left\{ \begin{tabular}{cc}
			$\frac{n!}{(n-k)!} a x^{n-k}$ & $k \leq n$ \\
			$0$ & $k>n$
		\end{tabular} \;\;\;.
		\right.
	\end{displaymath}
	
	Derivative of Equation (\ref{eq.2})
	
	Differentiation reduces the exponent by one:
	
	\begin{displaymath}
		\frac{d}{dx} x^a = ax^{a-1} \;\;\;,\;\;\; a = \frac{m}{n} \;\;\;.
	\end{displaymath}
	
	\noindent
	While $k$-th derivative is:
	
	\begin{displaymath}
		\dfrac{d^ky}{dx^k} = a \prod_{i=0}^{i=k} \Big(\dfrac{m}{n} - i \Big) x^{\dfrac{m}{n} - k} \;\;\;.
	\end{displaymath}
	
	\item The primitive of power function. 
	
	The primitive of Equation (\ref{eq.1}) is a power function of degree $n+1$:
	
	\begin{displaymath}
		\int_0^x (at^n)dt = \frac{a}{n+1} x^{n+1} \;\;\;.
	\end{displaymath}
	
	The primitive of Equation (\ref{eq.2}) for $a>-1$ it holds that
	
	\begin{displaymath}
		\int_0^x t^a dt = \frac{x^{a+1}}{a+1} \;\;\;,\;\;\; a=\frac{m}{n} \;\;\;,
	\end{displaymath}
	
	for $a<-1$ it holds that
	
	\begin{displaymath}
		\int_x^0 t^a dt = \frac{-x^{a+1}}{a+1} \;\;\;,
	\end{displaymath}
	
	for $a=-1$ it holds that
	
	\begin{displaymath}
		\int_1^x t^{-1} dt = \ln(x) \;\;\;.
	\end{displaymath}
	
\end{itemize}

Note that the power function is monotonically increasing, which is the main issue hindering the capacity to manage contrast in images adequately. Modeling tone is a resource that photographers use to manage exposure time. If a way exists to control the dichotomy in an acquired image already in a database or during the exposition time, the range of possibilities can dramatically change photography. 

\section{Dichotomy in Tone Image Analysis and Processing}
\label{Dichotomy}

The meaning of dichotomy is commonly and often confused. People generally use the term to talk about the division into two mutually exclusive or contradictory groups. Curiously, the word is found in the company of the term false, creating the meaning of a false dichotomy, which refers to the situation when someone is given only two choices when, in reality, there are many more available. The correct meaning of dichotomy proceeds etymologically speaking from the prefix "di," which means split into two or more, like in divide, which means to split into parts; dialogue implies a conversation between two or more people; and diverse, which, signifies varied or different—finally, the greek word "tomy" meaning cutting, incision. The following section will introduce a new mathematical model to manage tone and light to emphasize contrast. Before continuing with the mathematical analysis, let us introduce the concepts of contrast and tone.

Contrast refers to the relative brightness of any point in a color space, normalized to 0 for darkest black and 1 for lightest white. It is the difference in the luminance of a color that enables us to see the contrast. Regarding human perception, a difference in luminance matters more than a color difference. Tone, on the other hand, is simply a combination of color or a greyed-down hue. If a hue is mixed only with white or black, it stays a pure tint or shade. Hence, tone refers to the brightness levels in the photograph, from solid black to pure white. Shadows are dark tones, and highlights are bright tones. 

\begin{Definition}\label{image}{\bfseries[Image as the graph of a function]}.
	Let $f$ be a function $f:\mathrm{U} \subset \mathbb{R}^2 \rightarrow \mathbb{R}$. The graph or image $I$ of $f$ is the subset of $\mathbb{R}^3$ that consists of the points $(x,y,f(x,y))$, in which the ordered pair $(x,y)$ is a point in $\mathrm{U}$ and that $f(x,y)$ is the value at that point. Symbolically, the image $I=\{(x,y,f(x,y)) \in \mathbb{R}^3 | (x,y) \in \mathrm{U}\}$.
\end{Definition}

\begin{Definition}{\bfseries[Gamma correction]}\label{gamma_correction}
	Given an input image $I$ with values $f_{in}(x,y)$ and the transformed image $I'$ with output values $f_{out}(x,y)$. Usually, the relationship follows a power law.
	
	\begin{equation}
		f_{out}(x,y) = f_{in}(x,y)^\gamma \;\;\; 0 \leq f_{in} \leq 1 \;\;\;.
	\end{equation}
	
	\noindent
	The image increases the contrast in dark areas and loses it in light areas when $\gamma < 1$, while for $\gamma > 1$, the image rises its contrast in light areas and decreases it in dark places.
\end{Definition}

The result presents a difficulty in handling the dynamic range to adequately express the issue of tonal contrast. Note that the following lemmas associated to gamma correction depend on $\gamma$.

\begin{Lemma}\label{Lemma1}
	For any value of $\gamma \in (0, \infty)$, it holds that:
	
	\begin{equation}
		f_{out}(x,y) = 0^\gamma = 0 \ \ \ \ \text{and} \ \ \ \ f_{out}(x,y) = 1^\gamma = 1 \;\;\;.
	\end{equation}
\end{Lemma}

\begin{Lemma}
	Let the range of values for the exponent be $0 < \gamma < 1$, values of $f_{in}(x,y)$ close to $0$ result in output values that change rapidly, while for values $f_{in}(x,y)$ close to $1$; its change in the output values is small.
\end{Lemma}

\begin{Lemma}
	Let the range of values for the exponent be  $\gamma > 1$, values of $f_{in}(x,y)$ close to $0$ result in slowly changing output values, while for values $f_{in}(x,y)$ near $1$; its change in output values is significant.
\end{Lemma}

\begin{Lemma}
	For $\gamma = 0$, there are no changes in the output values.
	\begin{equation}
		f_{out}(x,y) = f_{in}(x,y)^0 = 1 \;\;\;.
	\end{equation}
\end{Lemma}

\begin{Lemma}\label{lemma5}
	For $\gamma \to \infty$, the values of $f_{in}(x,y)$ are defined by:
	\begin{equation}
		\lim_{\gamma \to \infty} f_{in}(x, y)^\gamma =
		\begin{cases}
			0 & 0 \leq f_{in}(x, y) < 1\\
			1 & f_{in}(x, y) = 1
		\end{cases}\;\;\;.
	\end{equation}
\end{Lemma}

\noindent
The properties of gamma correction are:

\begin{itemize}
	\item Periodicity: none
	\item Domain: $0 \leq f_{in}(x,y) \leq 1$
	\item Contradomain: $0 \leq f_{out}(x) \leq 1$
	\item Quadrant: First quadrant
	\item Monotonicity: Monotonically increasing
	\item Asymptotes: None
\end{itemize}

\begin{Theorem}{\bfseries[Dichotomy function]}.\label{theorem}
	Given an image $I$ and a transformation function $f_{out}(x,y)=f_{in}(x,y)^{\gamma_n}$, a reference function is calculated from the identity $\gamma_0 = 1$, in addition to a contrasted image which is obtained with a $\gamma_n \neq 1$. The term $f_{in}(x,y)^{1}=f_{in}(x,y)$ represents the identity, while $f_{in}(x,y)^{\gamma_n}$ are the $n$ different $\gamma \in \{\mathbb{R}^+ \cup 0\}$ corrections applied to the input value. The difference between the contrasted function and the reference function $f_{out}=k|f_{in}(x,y)^{\gamma_n}-f_{in}(x,y)|$ produces a function with a dichotomous behavior that magnifies the contrast produced by the transformation function, where $k$ is a scale factor to normalize the output between $0 \leq f_{out}(x,y) \leq 1$.. 
\end{Theorem}


\begin{proof}[Proof of Theorem 1]
	The application of the absolute value helps to calculate the actual distance from the reference function, although other approaches, such as the Euclidean distance, could be applied. Therefore, the resulting equation maximizes the difference of contrasts $DoC$ with two main slopes (positive and negative) and a unique inflection point where the maximum contrast located at $d_{\max}$ is $DoC_{\max}= \max(f(x,y)^{\gamma_{n}-1}) \forall n$.
\end{proof}
In the following we restrict the domain of definition. Properties of

\begin{equation}\label{doc}
	f(x)=|x^{\frac{m}{n}}-x| \;\;\; .
\end{equation}

\begin{tabular}{lll}
	Periodicity: & & None\\
	Domain of definition: & &	$0 \leq x \leq 1$ \\
	Range of values:	& &  $0 \leq f(x) < 1$ \\
	Quadrant: &  & first quadrant\\
	Monotonicity: & $x<d_{\max}$  & strictly monotonically increasing \\ 
	& $x>d_{\max}$: & strictly monotonically decreasing \\
	Asymptotes: & & None\\
\end{tabular}

\noindent
Remark 4. About the particular values, note that the function has two zeros at $x=\{0,1\}$, one pole at $x=0$, and there are no discontinuities. However, there is one extreme value which is a maximum at $x=d_{\max}$ and issues with the point of inflection or saddle points highlighted in the previous section but which are mimimized by the scale factor $k$.

In the following and for simplicity, the function applied to the image is given by:

\begin{equation}
	f_{out}(x, y; \gamma) = |f_{in}(x, y)^\gamma - f_{in}(x, y)| =
	\begin{cases}
		f_{in}(x,y)^\gamma - f_{in}(x,y) & 0 \leq \gamma \leq 1\\
		f_{in}(x,y) - f_{in}(x,y)^\gamma & \gamma > 1 
	\end{cases}\;\;\;.
\end{equation}

\noindent
The reciprocal of the image (see Definition \ref{image}) is:

\begin{equation}
	\dfrac{1}{f_{out}(x, y; \gamma)} =
	\begin{cases}
		\dfrac{1}{f_{in}(x,y)^\gamma - f_{in}(x,y)} & 0 \leq \gamma \leq 1\\
		\dfrac{1}{f_{in}(x,y) - f_{in}(x,y)^\gamma} & \gamma > 1 
	\end{cases}\;\;\;;
\end{equation}
\noindent
note that $f_{in}(x,y) \neq \{0,1\}$.

In the following all properties are given using Equation (\ref{doc}):

\noindent
Recursion relation for $\gamma = 2, 3, 4, ...$

\begin{equation}
	x - x^\gamma = x(1 - x^{\gamma - 1}) \;\;\;.
\end{equation}

\noindent
The sum of two polynomials, or the $\gamma$ root of the $n$ power, with $0 \leq \gamma \leq 1$:

\begin{equation}
	a(x^\gamma - x) \pm b(x^\gamma - x) = (a \pm b) (x^\gamma - x) \;\;\;,
\end{equation}

\noindent
with $ \gamma > 1$:

\begin{equation}
	a(x - x^\gamma) \pm b(x - x^\gamma) = (a \pm b) (x - x^\gamma) \;\;\;.
\end{equation}

\noindent
Product of two polynomials, or the $\gamma$ root of the $n$ power, with $0 \leq \gamma \leq 1$:

\begin{equation}
	a(x^\gamma - x)^c \times b(x^\gamma - x)^d = (ab) (x^\gamma - x)^{c+d} \;\;\;,
\end{equation}

\noindent
with $ \gamma > 1$:

\begin{equation}
	a(x - x^\gamma)^c \times b(x - x^\gamma)^d = (ab) (x - x^\gamma)^{c+d} \;\;\;.
\end{equation}

\noindent
The first derivative for $0 \leq \gamma \leq 1$:

\begin{equation}
	\dfrac{d}{dx} (x^\gamma - x) = \gamma x^{\gamma - 1} - 1 \;\;\;,
\end{equation}

\noindent
with $ \gamma > 1$:

\begin{equation}
	\dfrac{d}{dx} (x - x^\gamma) = 1 - \gamma x^{\gamma - 1} \;\;\;.
\end{equation}

\noindent
The $k$-th derivative for $0 \leq \gamma \leq 1$:


\begin{equation}
	\dfrac{d^ky}{dx^k} = 
	\begin{cases}
		\gamma x^{\gamma -1} - 1 & k = 1 \\
		\prod_{i=0}^{i=k} \Big(\gamma - i \Big) x^{\gamma - k} & k > 1
	\end{cases} \;\;\;,
\end{equation}

\noindent
with $ \gamma > 1$:


\begin{equation}
	\dfrac{d^ky}{dx^k} = 
	\begin{cases}
		1 - \gamma x^{\gamma -1} & k = 1 \\
		-\prod_{i=0}^{i=k} \Big(\gamma - i \Big) x^{\gamma - k} & k > 1
	\end{cases} \;\;\;.
\end{equation}

%
%
%

\noindent
The primitive with $0 \leq \gamma < 1$:

\begin{equation}
	\int_0^x (t^n - t)dt = \dfrac{1}{n+1}x^{n+1} - \dfrac{1}{2}x^2 \;\;\;,
\end{equation}

\noindent
with $\gamma > 1$:

\begin{equation}
	\int_0^x (t - t^n)dt = \dfrac{1}{2}x^2 - \dfrac{1}{n+1}x^{n+1} \;\;\;.
\end{equation}
\noindent           
Finally, the following lemmas and theorems result from Theorem \ref{theorem}.

\begin{Lemma}\label{Lemma6}
	According to the term $f_{out}(x, y) = f_{in}(x, y)^\gamma - f_{in}(x, y)$ when $\gamma = 0$, $f_{out}$ is the inverse of the image:
	
	\begin{equation}
		f_{out}(x, y) = f_{in}(x, y)^0 - f_{in}(x, y) = 1 - f_{in}(x, y) \;\;\;.
	\end{equation}
\end{Lemma}

\begin{Lemma}\label{Lemma7}
	According to the term $f_{out}(x, y) = f_{in}(x, y) - f_{in}(x, y)^\gamma$ when $\gamma = \infty$, $f_{out }$ behaves like the reference image $f_{in}$ except when $f_{in} = 1$
	
	\begin{equation}
		\lim_{\gamma \to \infty} (f_{in}(x, y) - f_{in}(x, y)^\gamma) =
		\begin{cases}
			f_{out} = f_{in} & 0 \leq f_{in} < 1\\
			f_{out} = 0 & f_{in} = 1 
		\end{cases}\;\;\;.
	\end{equation}
\end{Lemma}

\noindent
Note that lemma \ref{lemma5} helps to understand this result.

\begin{Lemma}\label{Lemma8}
	The transformation function $f_{out}$ has two roots in $f_{in} \in [0, 1]$. These roots are found at the values $f_{in} = 0$ and $f_{in} = 1$ for $\gamma \in \mathbb{R}^+$. The proof is in lemma \ref{Lemma1}.
\end{Lemma}

In the following all lemmas are given after Equation (\ref{doc}):

\begin{Lemma}\label{Lemma9}
	One local maximum or inflection point $d_{\max}$ exists for $0 < \gamma < \infty $ in the domain $x \in [0, 1]$.  Let $d_{\max}$ be the value of $f_{{\max}}(x) when f' = 0$, hence its slope is $0$ and lacks contrast. In the interval $x \in [0, d_{\max})$, the slope is always positive $f'(x)>0$, while in the domain $x \in (d_{\max}, 1]$ its slope is negative $f'(x)< 0$. Considering that $m = f'(x)$:
	
	\begin{equation}\label{eq.lemma9}
		m = 
		\begin{cases}
			m_+ & \dfrac{f(d_{max}) - f(0)}{d_{max} - 0} = \dfrac{f(d_{max})}{d_{max}}\\
			m_- & \dfrac{f(1) - f(d_{max})}{1 - d_{max}} = \dfrac{-f(d_{max})}{1 - d_{max}}
		\end{cases}\;\;\;.
	\end{equation}
	\noindent
	Considering $\gamma = 0$, $\gamma = \infty$, and Equation(\ref{eq.lemma9}) we have:
	
	\begin{equation}
		m =
		\begin{cases}
			\gamma = 0 & m_+ = \infty\\
			\gamma = \infty & m_- = \infty 
		\end{cases}\;\;\;.
	\end{equation}
	\noindent
	This is because when $\gamma=0$ then $d_{\max} = 0$ and $f(d_{\max}) = 1$ (given by Lemma \ref{Lemma6}), while at $\gamma = \infty$ the value $d_{max} = 1$ and $f(d_{\max}) = 1$ (given by Lemma \ref{Lemma7}).
\end{Lemma}

\begin{Lemma}\label{Lemma10}
	Since the function is highly nonlinear, the slope can be obtained pointwise (numerically) or in the region with the highest gradient magnitude ($m_+$ and $m_-$). Specifically, using finite differences for $0 < \gamma < 1$:
	
	\begin{equation}
		f'(x) = \lim_{\Delta x \to 0} \dfrac{(x+ \Delta x)^\gamma - x^\gamma - \Delta x}{\Delta x} \;\;\;,
	\end{equation}
	\noindent
	while for $\gamma > 1$:
	
	\begin{equation}
		f'(x) = \lim_{\Delta x \to 0} \dfrac{x^\gamma + \Delta x - (x+ \Delta x)^\gamma}{\Delta x} \;\;\;.
	\end{equation}
\end{Lemma}

\begin{Lemma}\label{Lemma11}
	Regionally and considering that the dichotomy function given by Theorem \ref{theorem} depends on $d_{\max}$, the area under the curve of $R_+$ and $R_-$ also depends on $d_{\max}$ implying that the smaller the value of its area, the greater its contrast. Hence, For $0 < \gamma < 1$, the positive region $R_+(x)$ is:
	
	\begin{equation}
		R_+(x) = \int_0^{d_{\max}} (x^\gamma - x)dx = \dfrac{d_{\max}^{\gamma + 1}}{\gamma + 1} - \dfrac{d_{\max}^2}{2} \;\;\;.
	\end{equation}
	\noindent
	And its negative region $R_-(x)$ is:
	
	\begin{equation}
		R_-(x) = \int_{d_{\max}}^1 (x^\gamma - x)dx = \dfrac{1 - d_{\max}^{\gamma + 1}}{\gamma + 1} + \dfrac{1}{2}(d_{\max}^2 - 1) \;\;\;.
	\end{equation}
	\noindent
	On the other hand, for values $\gamma > 1$, its positive region $R_+(x)$ is given by:
	
	\begin{equation}
		R_+(x) = \int_0^{d_{\max}} (x - x^\gamma)dx = \dfrac{d_{\max}^2}{2} - \dfrac{d_{\max}^{\gamma + 1}}{\gamma + 1} \;\;\;.
	\end{equation}
	\noindent
	Finally, its negative region $R_-(x)$ is defined by:
	
	\begin{equation}
		R_-(x) = \int_{d_{\max}}^1 (x - x^\gamma)dx = \dfrac{1}{2}(1 - d_{\max}^2) - \dfrac{1 - d_ {\max}^{\gamma + 1}}{\gamma + 1} \;\;\;.
	\end{equation}
\end{Lemma}

\begin{Lemma}\label{Lemma12}
	The area under the curve of function $f(x)$ for $\gamma \in [0, \infty)$ is less than or equal to Gamma correction evaluated in $\gamma_{0} = 1$ or the reference function in the image, hence $x^{\gamma=1}$: 
	
	\begin{equation}
		\int_0^1 f(x)dx = R_{+} + R_{-} \leq \int_0^1 xdx = 0.5 \;\;\;.
	\end{equation}
\end{Lemma}

\begin{Lemma}\label{Lemma13}
	According to Lemma \ref{Lemma6}, $\gamma = 0$ in the domain $x \in [0, 1]$, the slope $m_{-}$ is linearly negative with value $m_{-} = -1$, $R_{+} = 0$, and $R_{-} = 0.5$. Similarly after Lemma \ref{Lemma7} for $\gamma = \infty$ in the domain $x \in [0, 1)$ , the value of the slope $m_{+} = 1$, $R_{+} = 0.5$, and $R_{-} = 0$. Note that for $\gamma = 2$, the magnitudes of the positive regions $R_+$ and negative regions $R_-$ are symmetric at $x = 0.5$.
\end{Lemma}

\begin{Lemma}\label{Lemma14}
	The calculation of $k$ for all $\gamma > 0$ is defined as:
	
	\begin{equation}
		k = \dfrac{1}{\max( f(x; \gamma))} \;\;\;.
	\end{equation}
\end{Lemma}

Finally, an important property with practical applications of Theorem \ref{theorem} is the calculation of the inverse. However, the inverse function is difficult to obtain in general. Here, we propose two different methods based on the golden-section search for real numbers, and mapping the output function with look-up tables in the case of integers.

\begin{itemize}
	\item case 1: For all $\gamma \in (0,\infty)$, the values of $f_{in}$ can be recalculated from the values of $f_{out}$ even though the function is not invertible in many cases. Knowing $\gamma$, all values from $f_{out}$ can be mapped to $f_{in}$. However, the function has two roots at $0$ and $1$, and a local maximum $d_{\max} \in (0,1)$. Given the orientation of the slope $m$ it is possible to know in which region of the function $f_{in}$ belongs. Hence, any recursive numerical method can approximate the value $f_{in}$. Let $a=0$ and $b=d_{\max}$ be two extreme values within the region $C_+$, or let $a=d_{\max}$ and $b=1$ be two extreme values within the region $C_{-}$. Therefore, the goal is to minimize $f_{out} \leq \epsilon$ while using the function $f_{out}(x) = |x^\gamma - x|$, and the updated values $a, b$ are:
	
	\begin{equation}
		c = b - (b - a) / \varphi \ \ \text{and} \ \ d = a + (b - a) / \varphi \;\;\;,
	\end{equation}
	\noindent
	where $\varphi = \dfrac{1 + \sqrt{5}}{2}$; moreover, the approximation functions are given by:
	
	\begin{equation}
		|c^\gamma - c| - e \ \ \ \text{and} \ \ \ |d^\gamma - d| -e \;\;\;,
	\end{equation}
	\noindent
	if $f_{\max}(c) - e < f_{\max}(c) - e <$ then $b=d$, otherwise $a=c$. Note that the criterion ends when:
	
	\begin{equation}
		|f_{out}((b + a) / 2) - e| < \epsilon \;\;\;,
	\end{equation}
	
	\noindent
	where $\epsilon$ is a fixed error value. 
	
	\item Case 2: For the case when the image contains positive integer values (discrete values), a look-up table (LUT) can be used; this means that the value of $f_{in}(x,y)$ corresponds to the index of all calculated $f_{out}(x,y)$ values. To calculate $f_{in}$ from given $f_{out}$ and $\gamma$, and knowing that $f_{out}$ can have two equal values (considering $m_+$ or $m_-$) for each $f_{in}$, which is helpful to obtain the value of the index to which $f_{out}$ belongs.
	
	Given a finite list of positive integer values $f_{in}(l) \;\; l \in [0,b]$, where $b$ is the maximum known value, the LUT maps to a list of positive real values $f_{out}(l/b)$, the transformation from $f_{out}$ to $f_{in}$ is given by:
	
	\begin{equation}
		f_{in}(l) \to LUT \to f_{out}(l/b) \;\;\;.
	\end{equation}
	
	For its inverse transformation, that means that the transformed image with positive real values $f_{out}$ is inversely mapped to positive integer values $f_{in}$ as follows:

	\begin{equation}
		f_{in}(l) =
		\begin{cases}
			l = \arg(f_{out}(l/b)), l/b \in [0, d_{\max}] & l/b \leq d_{\max} 	\;\; \text{if}\;\; m_+\\
			l = \arg(f_{out}(l/b)), l/b \in [d_{\max}, 1] & l/b > d_{\max} \;\; \text{if}\;\; m_-
		\end{cases}\;\;\;.
	\end{equation}
	
\end{itemize}

		\section{Results}
		\label{Results}

In this section we present three examples to illustrate the model capacity from the viewpoint of tone and contrast. Since contrast, lightness, and color perception are affected by the context of the nearby colors and other features within a picture independently of the media. Predicting contrast is challenging and disposing of a mathematical relationship is key to emphasize the image content. Note that the proposed model does not corrupt the data only change the representation. Digital photography exposure relates to the amount of light that enters the sensor. A properly exposed photograph must accurately reflect the brightness levels into the sensor. Figure~\ref{fig1} shows an underexposed image that does not receive enough light. The picture is too dark. The details in the low light are invisible since the light was insufficient. The image is full of dark regions that almost reach pure black. Figure~\ref{fig2} shows an overexposed image that received too much light. The image is too bright, and details in the highlights are missing. Photographers say that light regions are burned because they contain only pure white. Note that pure white and black cases prevent us from recovering information. A photograph must constantly be correctly exposed to the shot.

Nevertheless, suppose it is impossible to have a correct exposure while shooting or because the image was already taken, like in a space mission. In that case, our analysis allows us to make image variations to enhance contrast information and correctly recover the information in the shoot despite exposition issues. Figure~\ref{fig3} presents a photograph combining underexposed and overexposed images. It is remarkable how well our model recovers information in the foreground without sacrificing that in the background. The results presented here aim to help us realize the importance of mastering tone to create impactful pieces that evoke strong emotions.

As we can appreciate, tone is a vital tool to produce contrast within a photograph, and the model introduced in this article helps create a sense of opposition and tension between different image regions to place our focus correctly on particular parts of the photograph.

In summary, the model improves contrast by adjusting and remapping image intensity values to the full display range of the data type. Appreciating an image with good contrast is the result achieved through the proposed model, as there are sharp differences between black and white, hence improving the dynamic range.

\subsection{Analysis of Theorem and Lemmas in the Three Typical Problems}

In the following, we relate the theorem and lemmas with the three photographs from two viewpoints. First, we explain the theorem and lemmas sequentially while relating them to all three figures. Second, we explain the theorem and lemmas for each photograph.

\begin{figure}[h]
\centering
\includegraphics[width=13.0cm]{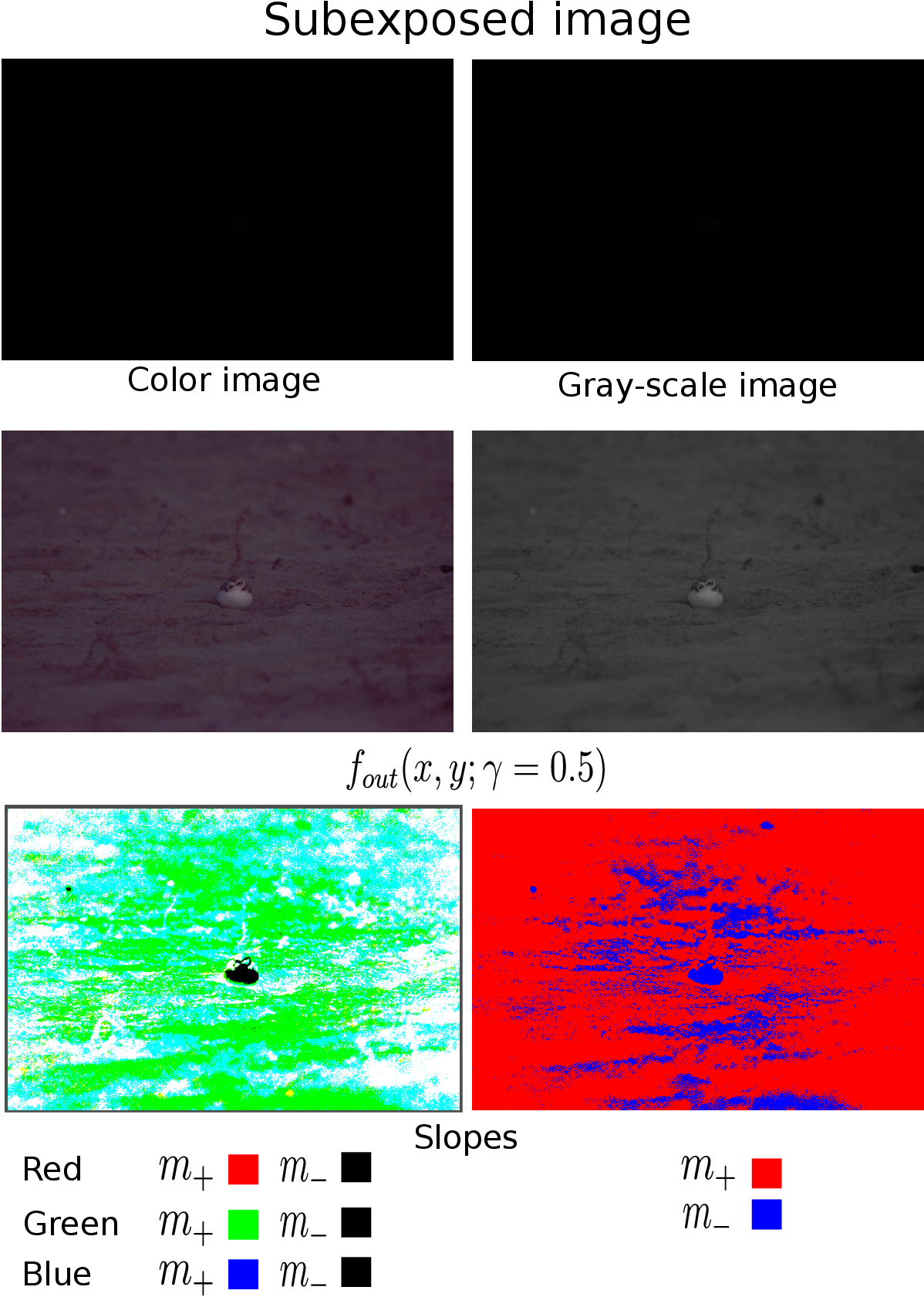}
\caption{Subexposed images are too-dark photographs with very little detail because the sensor receives insufficient light. Note that the proposed model correctly recovers tone dichotomy, allowing us to observe the snowy plover in its habitat. \label{fig1}}
\end{figure}

\begin{itemize}
	\item Theorem \ref{theorem} is visible in each figure's third row of images where $m_+$ and $m_-$ describe the type of region to which each pixel value belongs. For example, in the image of the third row and the second column, the mapping for each pixel is classified as red, which means that the pixel belongs to the class whose growth has a positive orientation ($m_+$). On the other hand, if the value is blue ($m_-$), the pixel belongs to the region with a negative orientation. Therefore, dichotomy appears since two different behaviors (given by their orientations) appear in the same image. The transformation helps to highlight the image display and extract information since the contrast is maximized for each region independently without changing the input information. Given the conditions of spatial illumination and naturalness of the input image, along with a correct gamma value, characteristics of the physics of light can be highlighted.
	An example is Figure \ref{fig3}; the image in the second row and first column shows a colored heron. Note the chromatic aberration that occurs at the edge of the bird at its extreme part: the external part of the body, neck, and head. On the other hand, in the adjacent figure, the same row, but in its grayscale version, the edges of the bird concerning its background are accentuated, allowing it to separate the object from the background in a more defined way.
	
	The value of theorem lies in the fact that the information contained in the image does not change geometrically but punctually since the calculations are given by the parameter $\gamma$, meaning that the function only works with the remainder between the input pixel values and the corrected pixel values (obtained after applying the gamma correction). Therefore, the resulting (contrasted) image is invertible by obtaining its original input values.
	
	Lemma \ref{Lemma6}, when $\gamma = 0$, the image has an inverse slope so that the entire image would correspond to its negative $f_{out}= 1 - f_{in}^\gamma$. The analysis did not include this particular case since it does not present a dichotomy but a change in the slope of the input values, although it is modelable in $f_{out}$. However, note that the $m_-$ regions of the three proposed figures contain regions with negative slopes; the only change is the slope's magnitude.
	
	Lemma \ref{Lemma7}, when $\gamma = \infty$, represents the opposite case of Lemma 6. The input values stay still since the values after applying the power function tend to zero, so the output image would correspond to the input image except for pure white since these values will change to pure black.
	
	Lemma \ref{Lemma8}, considering that there is a local maximum within the luminance range, the value of the function is 0 when it is pure white or pure black. None of the three images in the figures contain pure blacks or pure whites since there is no value to transform, just like the traditional gamma correction. However, a particular case occurs with the binary input images, whose output would be 0 for any $\gamma \in (0, \infty)$. For image handling, $\gamma = 0$ can be used by inverting the values in the negative of the input image.
	
	Lemma \ref{Lemma9} directly applies in the three figures ($m_+$ and $m_-$), whose simplest case is the grayscale version. The red areas have a positive slope, contrasting the image ascendingly. On the other hand, in the blue regions, their slope is negative; this means that the transformation would be equivalent to the image's negative. When the image is in color, the transformation applied to each color channel produces a similar behavior. In the red channel, $m_+$ is positive and ascending while $m_-$ is negative and descending, and similarly for the green and blue channels. When $m_+$ is yellow, the red and green channels are ascending, and the blue is descending, while magenta $m_+$ corresponds to red and blue ascending, and $m_-$ corresponds to green descending. When the color is cyan, the green and blue channels are ascending, and the red is descending, while the white color means that it is ascending in the three channels and black means that all the channels are descending. Note that the negative slope is related to lemma \ref{Lemma6}, and the positive slope is related to lemma \ref{Lemma7}. The differences in values at the edges can be accentuated or diminished with the gamma value for images that are underexposed, overexposed, or both cases.
	
	Considering Lemma \ref{Lemma10}, while it is difficult to approximate an appropriate gamma value to contrast a given image, one can numerically calculate the slope values $m_+$ or $m_-$ for a given gamma value. For example, in Figure \ref{fig3}, the overexposed region (image background) pixel values are very different from the heron, while the bird has similar values considering an underexposed region. Using a gamma value such that $DoC_{max}$ is between both regions will easily allow the contrast of both areas.
	
	Considering Lemma \ref{Lemma11}, and since the objective is to contrast the information content of the image, it is required to increase the magnitude of the slope, either positive or negative, with the input slope (the reference image has $m=1$). The local maximum is found in the range [0, 1) in $\gamma \in [0, \infty)$ and depends on the image exposure type as well as the magnitude of the slope. For example, for figure \ref{fig1}, a $\gamma=0.5$ was used, meaning that $d_{\max} = 0.25$, $f_{out}(0.25) = 0.25$ (corresponding to a pixel value with a depth of $8$ bits of $67.5$), and a $k = 1/0.25 = 4$ (Lemma \ref{Lemma14}). Under these parameters, it is possible to verify that calculating the integrals $R_+$ and $R_-$ produces that region $R_+$ will grow faster than $R_-$; in other words, $m_+$ has a greater magnitude than $m_-$.
	
	Regarding Lemmas \ref{Lemma12} and \ref{Lemma13}, for any image containing integer or real pixel values, the output $f_{out}$ will have the same range as $f_{in}$ for any positive gamma value, including $0$. As is well known, power functions are non-linear; however, $f_{out}$ has symmetric behavior when $\gamma = 2$. This means that $d_{\max} = 0.5$ (corresponding to the value $128$ in pixels with depth of $8$ bits), $DoC_{\max} = 0.25$ and $k = 1 / 0.25 = 4$. In this case, $m_+ = m_-$ and $R_+ = R_-$.
	
	Finally, taking into account lemma \ref{Lemma14} to normalize the function $f_{out}$, $k$ is calculated; this means that $k * DoC_{\max} = 1$, thus the function $f_{out}$ is scaled so that $f_{out}= k|x^\gamma - x|$. Once normalized, it is necessary to scale the result by the bit depth (usually $8$ bits) for future processing. The lemma \ref{Lemma14} and its two cases are helpful to transform $f_{out} \rightarrow f_{in}$. Any image must be normalized and scaled to the desired bit depth for visualization and storage in a standardized format.
	
	\item This section organizes each figure as follows. The first row is a pair of input images, on the left is the color image, and on the right is its grayscale version. The second row represents the transformed images with $f_{out}$ for a given $\gamma$ value. The third row contains images showing the slopes (and regions) to which each image pixel value belongs. For any of the figures, Lemma \ref{Lemma8} holds, since when the input values $f_{in}$ are pure white or pure black, the output $f_{out}$ will be pure black; this is because they do not provide information at the pixel level. However, Lemmas \ref{Lemma9} and \ref{Lemma11} are used in conjunction with the histogram to choose an appropriate value of $\gamma$ to subsequently calculate $k$ using Lemma \ref{Lemma14}. Knowing the distribution of the input pixel values, Lemma \ref{Lemma9} can be used to control the magnitude of the slope. In contrast, Lemma \ref{Lemma11} allows one to decide what relationship exists between the slope for a given $\gamma$.
	
	In Figure \ref{fig1}, the original image, either in color or grayscale, contains shallow pixel values, so a $\gamma$ value close to 0 would be ideal to increase the pixel values. Since the pixel values are compressed and close to $0$, it is necessary to decompress the image; a $\gamma$ value close to 0 fits. For $\gamma=0.5$, $d_{max} = 0.25$, $f_{out}(0.25) = 0.25$ and $k=4$, using Lemma \ref{Lemma9}, we have that $m_+=1$, while $m_- = -0.333$, these values multiplied by the scaling factor gives $k \times m_+ = 4$ and $k \times m_- = -1.333$. Note that even the magnitude of $k \times m_-$ is larger than the magnitude of the slope of the original image ($\gamma=1 \to m = 1$). This is also verifiable with Lemma \ref{Lemma11}, $R_+ = 0.11458$, while $R_- = 0.11458$. With the value of the regions, one can verify Lemma \ref{Lemma12}; $R_+ + R_- = 0.22916 < 0.5$. The third row of Figure \ref{fig1} gives a more graphical explanation of the dichotomy behavior. For example, for the grayscale version, the red mapped pixels belong to $m_+$ and $R_+$; their values contrast at a higher rate than the blue mapped pixels with a lower slope magnitude $m_-$. The dichotomy is present since both regions coexist in the same image. In the case of color, the situation is similar. Note that the pixels are black because their slopes are negative. Green pixels have a positive slope in the green channel, cyan pixels have a positive slope in the green and blue channels and a negative slope in the red channel, while white pixels have a positive slope in all three channels.
	
	Figure \ref{fig2} is the opposite case to Figure \ref{fig1}. The image is overexposed, so pixels throughout the image tend to be pure white. Therefore, any value of $\gamma$ is helpful for its transformation; for example, $\gamma = 1.2$. The transformation allows a contrast image by decompressing the pixel values using the histogram. With the value $\gamma = 1.2$, then $d_{max} = 0.41$, $f_{out} = 0.066963266$ and $k = 14.93356$. Using Lemma \ref{Lemma9}, we have that $m_+=0.163325$, $m_- = -0.113496$, so their scaled magnitude would be $k \times m_+ = 2.43935$ and $k \times m_- = -1.694899$. Starting from Lemma \ref{Lemma11}, the region $R_+ = 0.020121$, while $R_- = 0.025335$. With the regions, we can verify Lemma \ref{Lemma12}; $R_+ + R_- = 0.045456 < 0.5$. As in Figure \ref{fig1}, the third row contains the mapping of the slopes of each pixel. Graphically, the dichotomy appears in the non-chaotic transformation of the image pixels. Note that both Figures \ref{fig1} and \ref{fig2} can be improved by the original gamma correction, for example, with $\gamma=0.2$ for Figure \ref{fig1} and $\gamma=4$ for Figure \ref{fig2}. Still, the image in Figure \ref{fig1} comes out very bright, lacking contrast compared to our method, while in Figure \ref{fig2}, both methods perform similarly. Our method works with the residual between the gamma-corrected image and the original image instead of working only with the original image, hence the best results are when the image has both exposure issues.
	
	Figure \ref{fig3} is a combination of two regions, underexposed and overexposed. The image is of a heron, but the camera captures the photo poorly because the photographer selected the camera parameters to capture the intensity of the background instead of the object. As the background is brighter, the camera adjusts to the background region, so the object becomes darker, losing prominence in the composition. To enhance the image, we apply a value $\gamma=1.8$; therefore, $d_{\max} = 0.48$ and $k=4.691085$. According to Lemma \ref{Lemma9}, we have $m_+ = 0.444104$, $m_- = -0.4099428$, their scaled magnitudes $k \times m_+ = 2.0833$, $k \times m_- = -1.92307$ (note that the slopes converge until they become equal, see Lemma \ref{Lemma13}). Considering Lemma \ref{Lemma12}, and knowing the size of their regions $R_+ = 0.069458$ and $R_- = 0.0734$, we can verify Lemma \ref{Lemma12}; $R_+ + R_- = 0.142858 < 0.5$. Then, it is possible to appreciate the Theorem \ref{theorem} in the third row due to the mapping of slopes computed from the transformed image (red, green, cyan, yellow, black, and white for the color and finally, the grayscale version) as well as in the transformed image itself, which correspond to the second row. Moreover, it is possible to appreciate details in previously underexposed or overexposed areas. Note also in the color image that the edges, mainly on the contour object highlighted from the background, are much more defined, allowing the appreciation of chromatic aberrations (purple fringing) due to the environment's conditions and the lens's defect. On the other hand, considering the transformed image into a grayscale, the contours stand out from the background and object while improving other aspects, such as feature extraction.
	
	The transformation $f_{out}$ does not change the image's geometry; therefore, its information is invertible, and other image processing methods can be applied—for example, convolution, morphological processing, or preprocessing for higher-level computer vision tasks.

\end{itemize}

\begin{figure}[h]
\centering
\includegraphics[width=13.0cm]{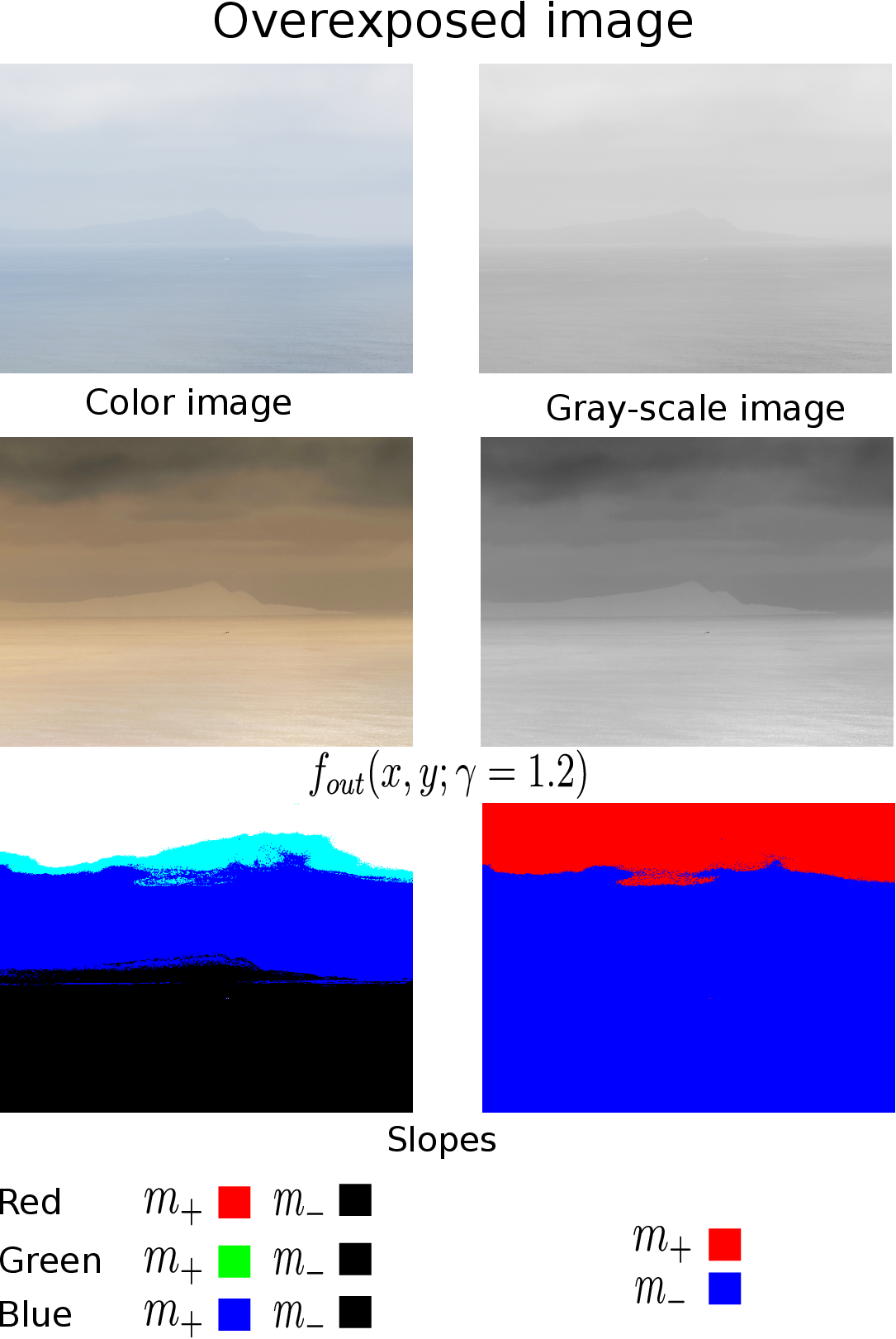}
\caption{A photograph is overexposed when the sensor receives too much light. Consequently, the picture is too bright, and details in the highlights are lost. Note that the model can create a correctly contrasted image feeling just bright or dark enough. \label{fig2}}
\end{figure}  
		
\begin{figure}[h]
\centering
\includegraphics[width=13.0cm]{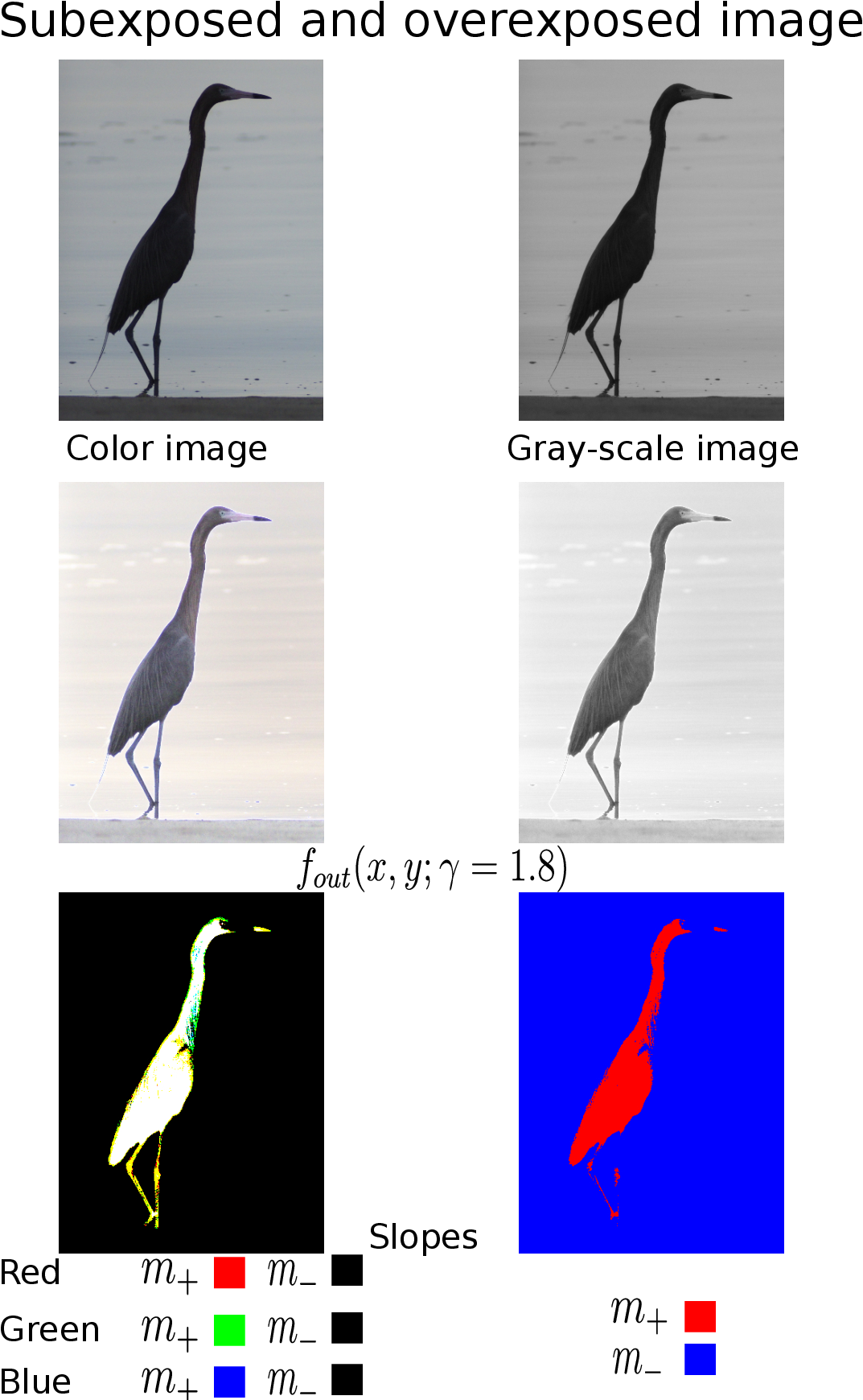}
\caption{The exposure is often problematic, and the image contains a mixture of underexposed and overexposed regions. Note that the transformation recovers the heron details, strikingly balancing the shadows and highlights. \label{fig3}}
\end{figure}  

\subsection{Quantitative and Qualitative Analysis of Dichotomy}

In this section, we will quantitatively and qualitatively (visually) analyze the dichotomy function, the traditional gamma correction, and a widely used method in the community based on calculating the gamma value using the histogram distribution function of the image. Note that we took photos with different cameras. In Figure \ref{fig1} (underexposed image), we used a Nikon D7200 camera, 500 mm focal length lens with f5.6 aperture at a resolution of $6000 \times 4000$. Figures \ref{fig2} and \ref{fig3} show two photographs (overexposed and mixed case) we took with an Olympus OMD-EM10 Mark III camera. These two images have Vivitar lenses 200 mm with f5.6 aperture and Brightin Star 50 mm with f8 aperture at resolutions $4608 \times 3456$ and $2876 \times 3915$ pixels, respectively. The underexposed and mixed cases use long focal lengths and open apertures, causing a reduction in the depth of field so that what is in front of and behind the focus area is blurred. As a result, the birds look closer and sharper. On the other hand, in the overexposed case, with a short focal length and closed aperture, the image is in focus, and the mountain appears far away. For the dichotomous function, the gamma values will be $\gamma = \{0.35, 1.2, 1.8\}$ and the traditional gamma correction $\gamma = \{0.25, 0.8, 4.0\}$ (sorted by the underexposed, overexposed and mixed cases). These gamma values were used for each figure since they presented better contrast (visually) overall throughout the image.

\begin{figure}[h]
	\centering
	\includegraphics[width=14.0cm]{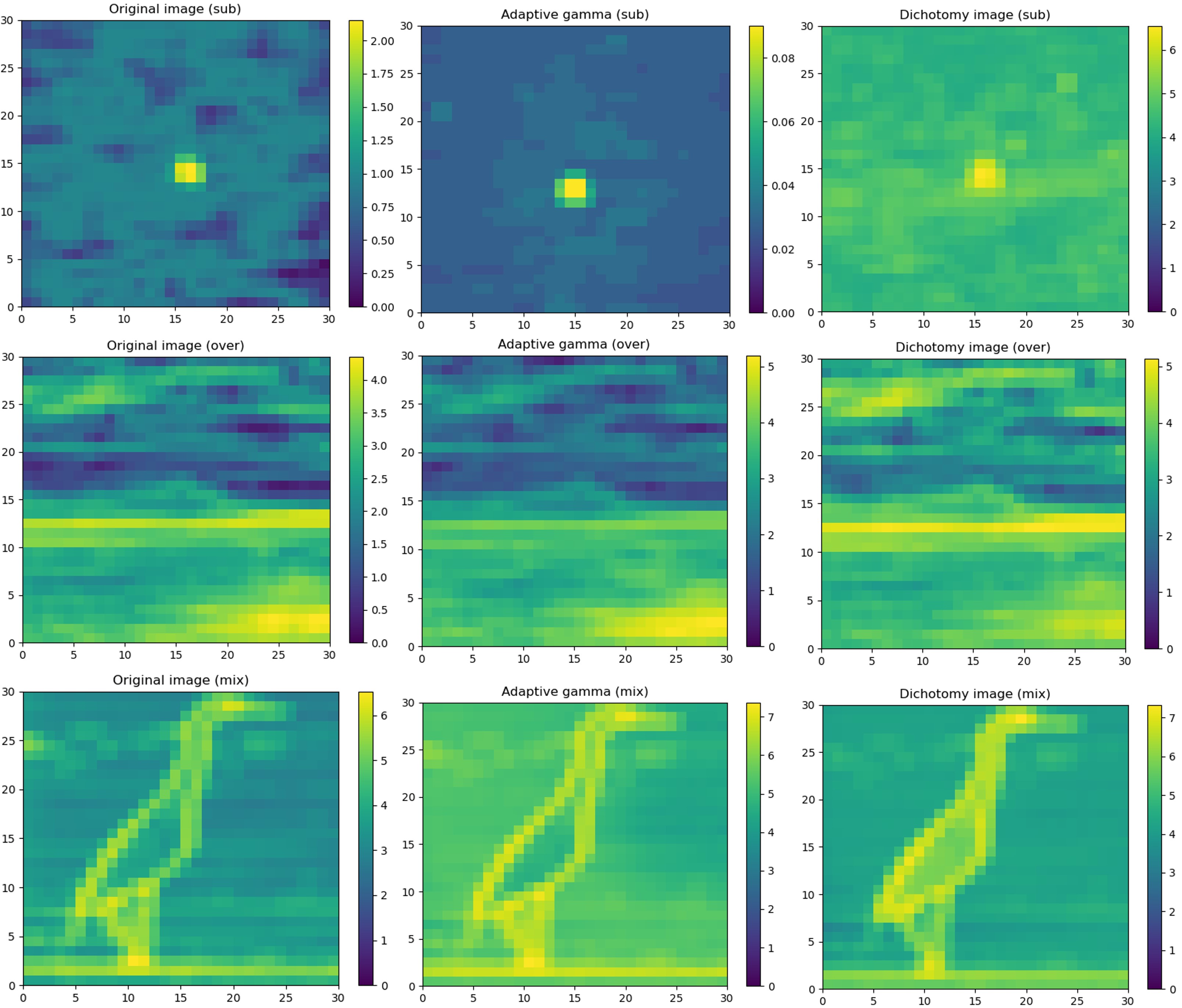}
	\caption{This collage shows the results of computing the entropy of Figures \ref{fig1}, \ref{fig2}, and \ref{fig3}, considering the original image, and after applying adaptive gamma and dichotomy functions. \label{fig7}}
\end{figure}

\begin{figure}[h]
	\centering
	\includegraphics[width=14.0cm]{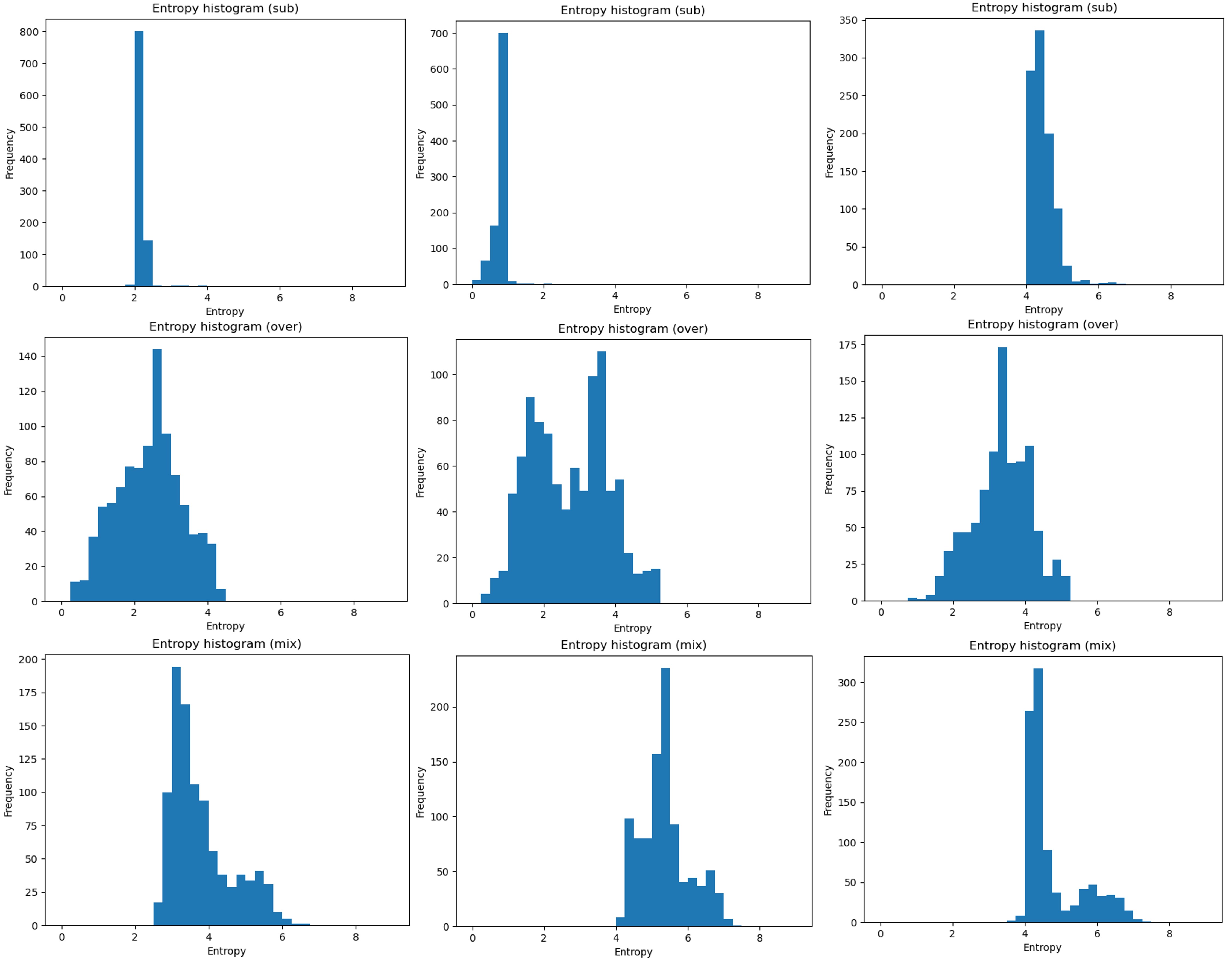}
	\caption{These charts provide the histograms computed from the results of Figure \ref{fig7}. \label{fig8}}
\end{figure}

The first evaluation considers entropy locally (patch extraction); our reason for local evaluation is that using the histogram globally occludes information about the contrast in regions with little variation due to the light condition (underexposure, overexposure, or mixed). Also, noise is part of the sampling, and it is observable when the images are high-resolution, with lossless compression, or have low illumination (consider that noise is more visible when illumination is low and exposure time is short). This evaluation aims to know the randomness of pixel values at a global level, given that if the entropy is low, then the patch has little contrast, and you can extract little information. On the contrary, if the variation is high, it is more likely to extract information that may not be relevant. In \cite{focus}, authors use entropy to know the image focus quality, and we adopt such a procedure to know the image variations. Consider a mesh of $30 \times 30$ patches; the patch size is calculated from the image size, so the patches do not keep the same size. We compute the entropy $H$ from the pixel values as follows:

\begin{equation}
	H = -\sum_{k=0}^{255} p(k)log_2(p(k)) \;\;\;,
\end{equation}

\noindent
with $p(k)$ approximated by the histogram of pixel intensity values taken from the patches using $k$ values. Figure \ref{fig7} shows the results of the computed entropy organized by row for the cases underexposed, overexposed, and mixed, and by columns are the original image, adaptive gamma, and the dichotomous function. Figure \ref{fig8} has the same organization, but the charts provide the histograms of the calculated entropy of Figure \ref{fig7}. For the first row, the underexposed case, both original image and adaptive gamma, have very little entropy, because the pixel values within the patches are very similar while the dichotomous function stretch the pixel values more comfortably. For the overexposed case, the original image presents less entropy; most patches range between 2 and 3, while adaptive gamma improves the range between patches. However, many patches have entropy values lower than four. For the dichotomy function, the entropy values of most patches concentrated around 3.5 and above; therefore, increasing the probability of extracting quality information. In the mixed case, the original image has slight background and bird body variation, but greater variation in the image contour. In contrast, the adaptive gamma has a more significant variation in the background than both the original image and the dichotomy function. Still, the dichotomous function has a more considerable variation in the bird body than adaptive gamma. For all 3 cases, the dichotomous function not only increases the variation of all patches but also makes the entropy of the patches homogeneous, which equalizes the spatial information throughout the image.

The following test visualizes contour extraction using adaptive gamma, gamma correction, and dichotomous function. We consider these three algorithms because such algorithms can handle images of any resolution (unlike neural networks), and the resulting images can be invertible to the original image. Although contour extraction has been widely studied and used in computer vision \cite{survey}, it still presents certain problems such as lighting or resolution issues, therefore our visual evaluation provides more information about the scope of contour extraction. We present two difficulties: first, the image's resolution is huge, greatly reducing the edge values, and second, the noise is more prominent in the image. To combat this, we present the deployment of an illumination space similar to the scale space \cite{scalespace} to learn the detector's behavior and progressively decrease the noise level of the image. This space first uses Gaussian smoothing with $\sigma^2 = 1, 2, 3, \ldots$ and then downsampling the image by $\sigma^2$. Second, due to the resolution, it is difficult to parameterize a detector since the size and thickness of the contours not only vary with the image size (inner scale) but also with the smoothing size (outer scale) \cite{innerouter}. For example, a detector with a small smoothing level will highlight small edges, thus making noise visible. Still, it will attenuate the saliency of thick contours, while a large smoothing value will do the opposite. To overcome this problem, we propose using a difference-of-Gaussians (DoG) detector with specific values
suitable for the image size of the scale space. The DoG \cite{scalespace} edge detector as an approximation of the Laplacian is given by:

\begin{equation}
	\dfrac{DOG(x; t, \Delta t)}{\Delta t} = \dfrac{L(x; t + \Delta t) - L(x; t)}{\Delta t}
\end{equation}

\noindent
with $\Delta t = (s^2 - 1)t$ due to a self-similar scale sampling $\sigma_{i+1} = s \sigma_{i}$ corresponding to $t_{i+1} = s^2 t_i$, $s$ is a tuning parameter, and 

\begin{equation}
	L(x, y; t) = \int_{(u,v) \in \mathbf{R}^2} f(x - u, y - v)g(u,v; t)dudv ,
\end{equation}

\noindent
define a Gaussian scale-space representation $L: \mathbf{R}^2 \times \mathbf{R_+} \to \mathbf{R}$ with $f: \mathbf{R}^2 \to \mathbf{R}$ being a two-dimensional image and $g:\mathbf{R}^2 \times \mathbf{R}^+ \to \mathbf{R}$ denotes the (rotationally symmetric) Gaussian kernel:

\begin{equation}
	g(x, y; t) = \dfrac{1}{2 \pi t}e^{-(x^2+y^2)/2t},
\end{equation}

\noindent
with variance $t=\sigma^2$. Therefore, a dichotomy space can be generated by varying the gamma value through Theorem \ref{theorem} using DOG:

\begin{equation}
	\frac{DOG_{dichotomy}(x; \gamma,t, \Delta t)}{\Delta t} = k \dfrac{|L(x; t + \Delta t)^{\gamma} - L(x; t + \Delta t)| - |L(x; t)^{\gamma} - L(x; t)|}{\Delta t}
	,
\end{equation}

\noindent
while Definition \ref{gamma_correction} helps building a gamma correction space with DOG: 

\begin{equation}
	\dfrac{DOG_{\gamma}(x; \gamma, t, \Delta t)}{\Delta t} = \dfrac{L(x; t + \Delta t)^{\gamma} - L(x; t)^{\gamma}}{\Delta t}.
\end{equation}

Hence, $L_{dichotomy}$ and $L_{\gamma}$ are spaces organized by the gamma values
$\gamma$ which means that each dichotomy function applied to DOG can highlight different types of pixels by the compression or decompression factor of the gamma value.

\begin{figure}[h]
	\centering
	\includegraphics[width=13.0cm]{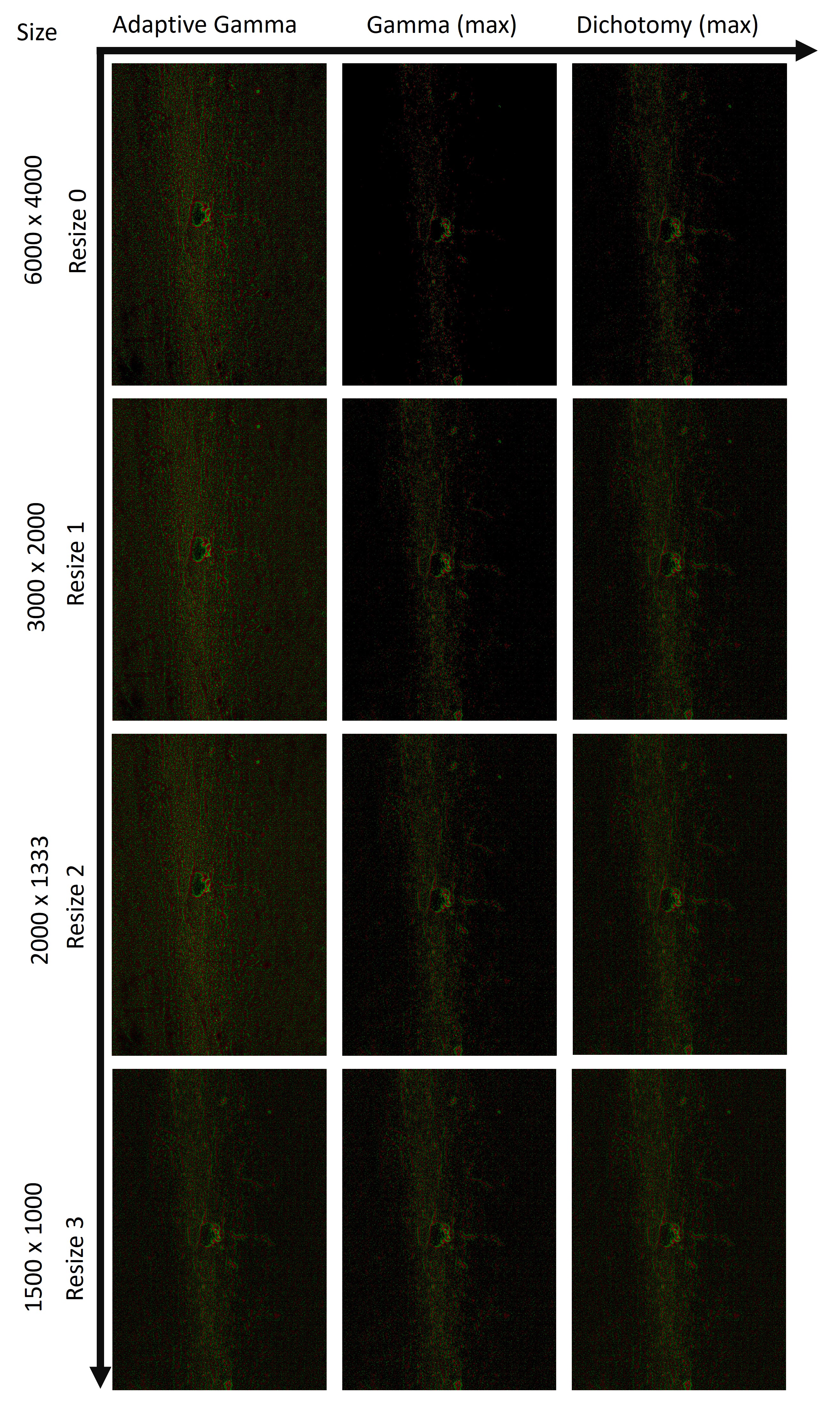}
	\caption{This collage shows . \label{fig9}}
\end{figure}

\begin{figure}[h]
	\centering
	\includegraphics[width=13.0cm]{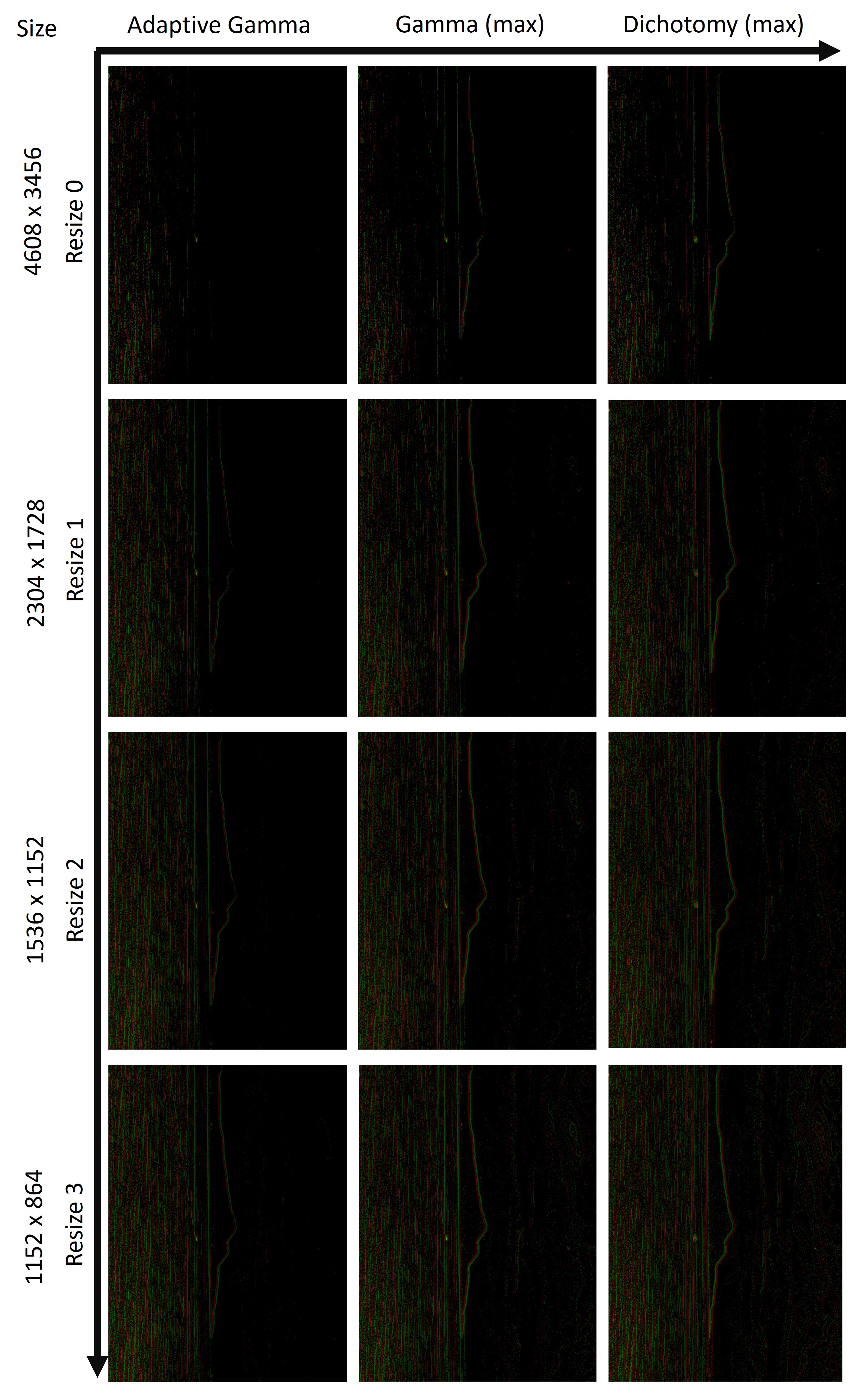}
	\caption{This collage shows . \label{fig10}}
\end{figure}

\begin{figure}[h]
	\centering
	\includegraphics[width=13.0cm]{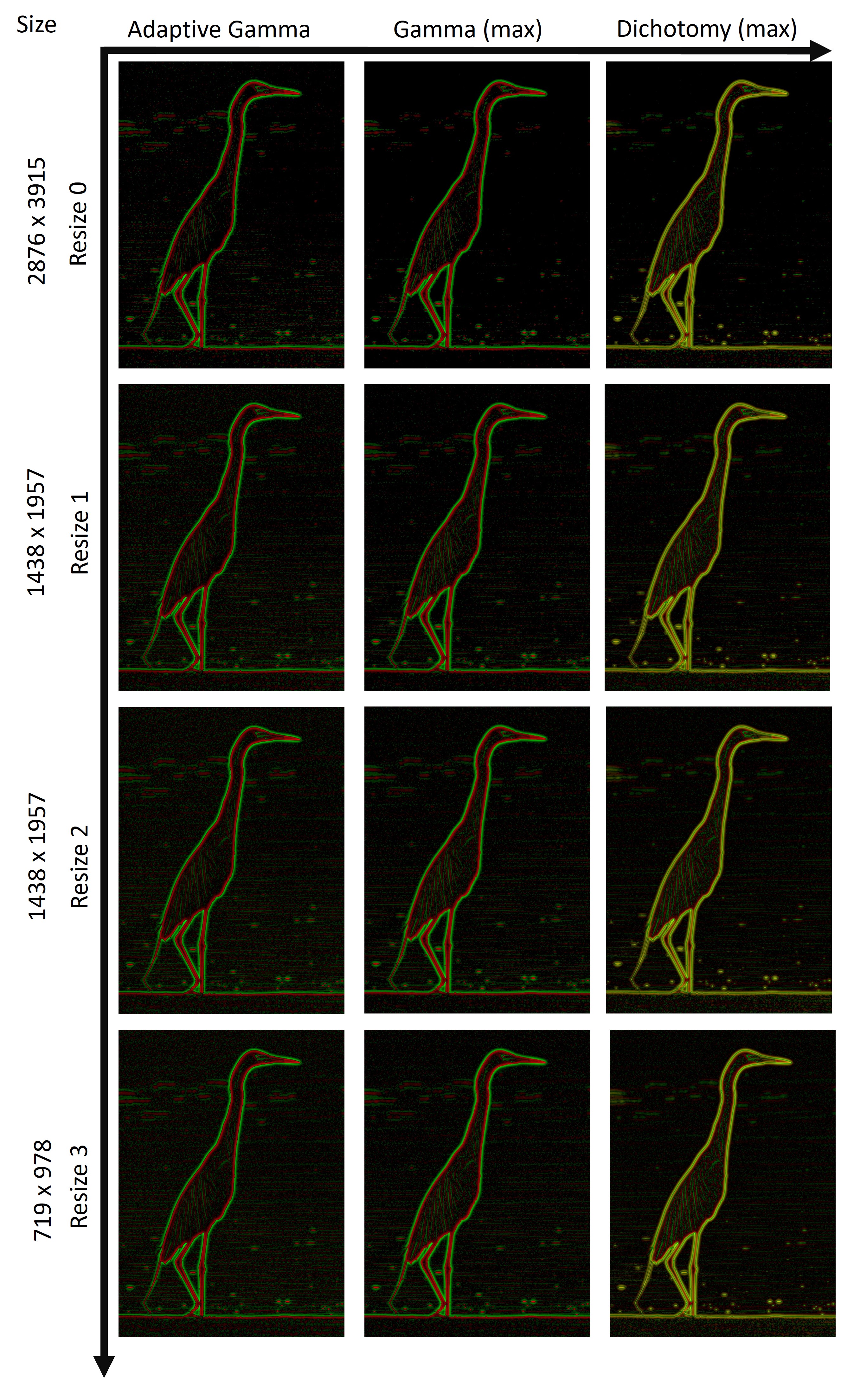}
	\caption{This collage shows . \label{fig11}}
\end{figure}

Once we obtain DOG, the values close to zero are filtered out, keeping only the local extremes. Note that we can discriminate between both maxima and for visualization purposes, we colored positive values with red and negative values with green, thus highlighting the local maxima and minima, see Figures \ref{fig9}, \ref{fig10}, and \ref{fig11}. We can extend this classification to all gamma applied in the $L_{dichotomy}$ like in the graphical abstract, where each color highlights DOGs applied with different gammas.
For $DOG_{dichotomy}(x,y; t)$:

\begin{equation} DOG_{d_{max}}(x, y) = \max_{\gamma}(DOG_{dichotomy}(x,y; \gamma, t, \Delta t)) , \end{equation} 

and for the negative: 

\begin{equation} DOG_{d_{min}}(x, y) = -\min_{\gamma}(DOG_{dichotomy}(x,y; \gamma, t , \Delta t)) , \end{equation} 

while for the gamma correction space $DOG_{\gamma}(x,y; t)$ 

\begin{equation} DOG_{\gamma_{max}}(x, y) = \max_{\gamma }(DOG_{\gamma}(x,y; \gamma, t, \Delta t)) , \end{equation} 

and for the negative: 

\begin{equation} DOG_{\gamma_{min}}(x, y) = -\min_{\gamma}(DOG_{\gamma}(x,y; \gamma, t, \Delta t)). \end{equation}

For the experimental stage, we apply a resize to the original image with values: $\sigma^2 = \{1, 2, 3, 4\}$, whose subsamples 1 and 3 are octaves of the image and 2 correspond to the dimensions divided by three, see Figure \ref{fig9}, \ref{fig10}, and \ref{fig11}. We enhance each subsampled image with $\gamma = \{0.25, 0.38, 0.5, 1.2, 1.8, 2, 2.4, 4\}$. Since DOG is an approximation of the normalized Laplacian, we compute DOG with those eight values, $t=8.192/ \sigma^2$, and $\Delta t=1$, while excluding positive and negative local extrema values with the following threshold values: $thr_+ = 0.2$ and $thr_- = -0.2$. 

We organized the three figures by columns representing the algorithms (adaptive gamma, gamma correction space, and dichotomy space) and the rows showing the pyramid-type scale space. Figure \ref{fig9}, underexposed case, the algorithm that extracted the most information in the first row was adaptive gamma, followed by the dichotomous and gamma spaces. The fact that we observe more signals may be due to two factors. The first is the noise present, evident since the out-of-focus area is far from the bird. Therefore, the front and back area of the bird is fuzzy and indistinct, which presumes that it is noise. The second factor arises from the difference between $t$ and $\Delta t$; if the detected objects are small, corresponding to blurred regions, most probably noise, especially when the image is large. 

In summary, the extracted information is within focus for the dichotomous space and extracts more information than the gamma space. As the image is subsampled, more information appears in the gamma and dichotomy spaces, whereas adaptive gamma begins to detect fewer contours. This phenomenon supports factor two since the information concentrates on smaller scaled-images.

Figure \ref{fig10}, the overexposed case, shows that the dichotomous space extracts more information in the entire scale than the gamma and adaptive gamma spaces. Note that even from the second subsampling, it extracts the contours of the clouds, the mountain's silhouette is well-defined, the sea appears contoured, and it does not seem to have traces of noise. Note also that the dichotomous space's local extremes (red and green) appear inverted in contrast to gamma and adaptive gamma spaces since, in this particular region, the data and the dichotomy function produce $m_-$ calculating the negative slope.

For Figure \ref{fig11}, the mixed case, the dichotomy space extracts much more information in the dark area and a bit of information in the light areas for the entire scale space. Likewise, the gamma space extracts more information than the adaptive gamma because gamma is a space representing more information than a single gamma. Note that the bird's outline in the dichotomous space is yellow; this is because the algorithm detected those contours in different gamma values, making it more robust to lighting changes.

\subsection{Comparison with Image Enhancement Approaches}

\begin{figure}[h]
	\centering
	\includegraphics[width=14.0cm]{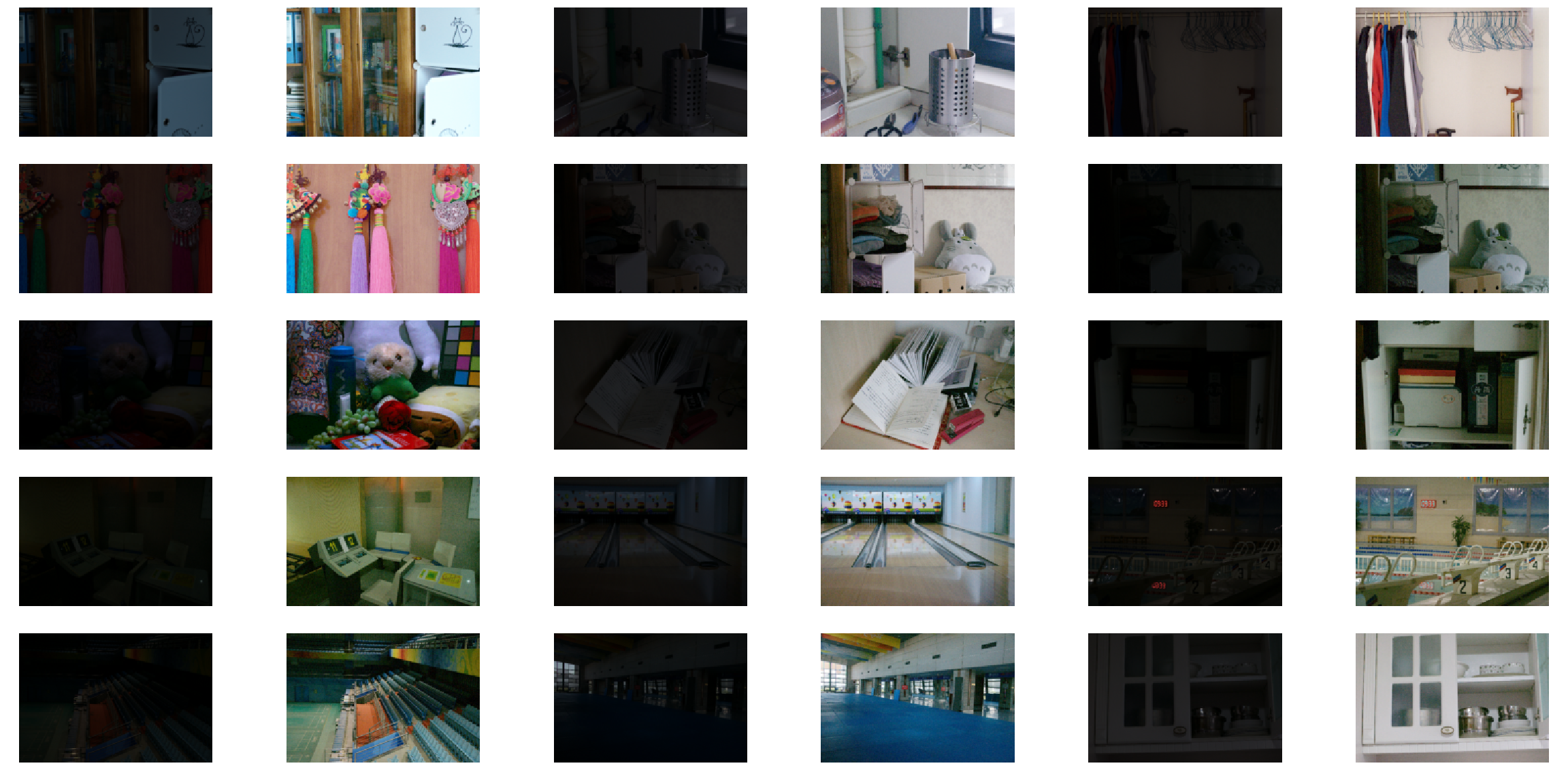}
	\caption{This collage shows the fifteen low-light testing images in the LOLv1 database, and the photos improved with the dichotomy model presented in this work. Surprisingly, it reproduces tones like daytime, with a simple mathematical operation that contrasts state-of-the-art models based on complex reasoning. \label{fig4}}
\end{figure}  

\begin{figure}[h]
	\centering
	\includegraphics[width=14.0cm]{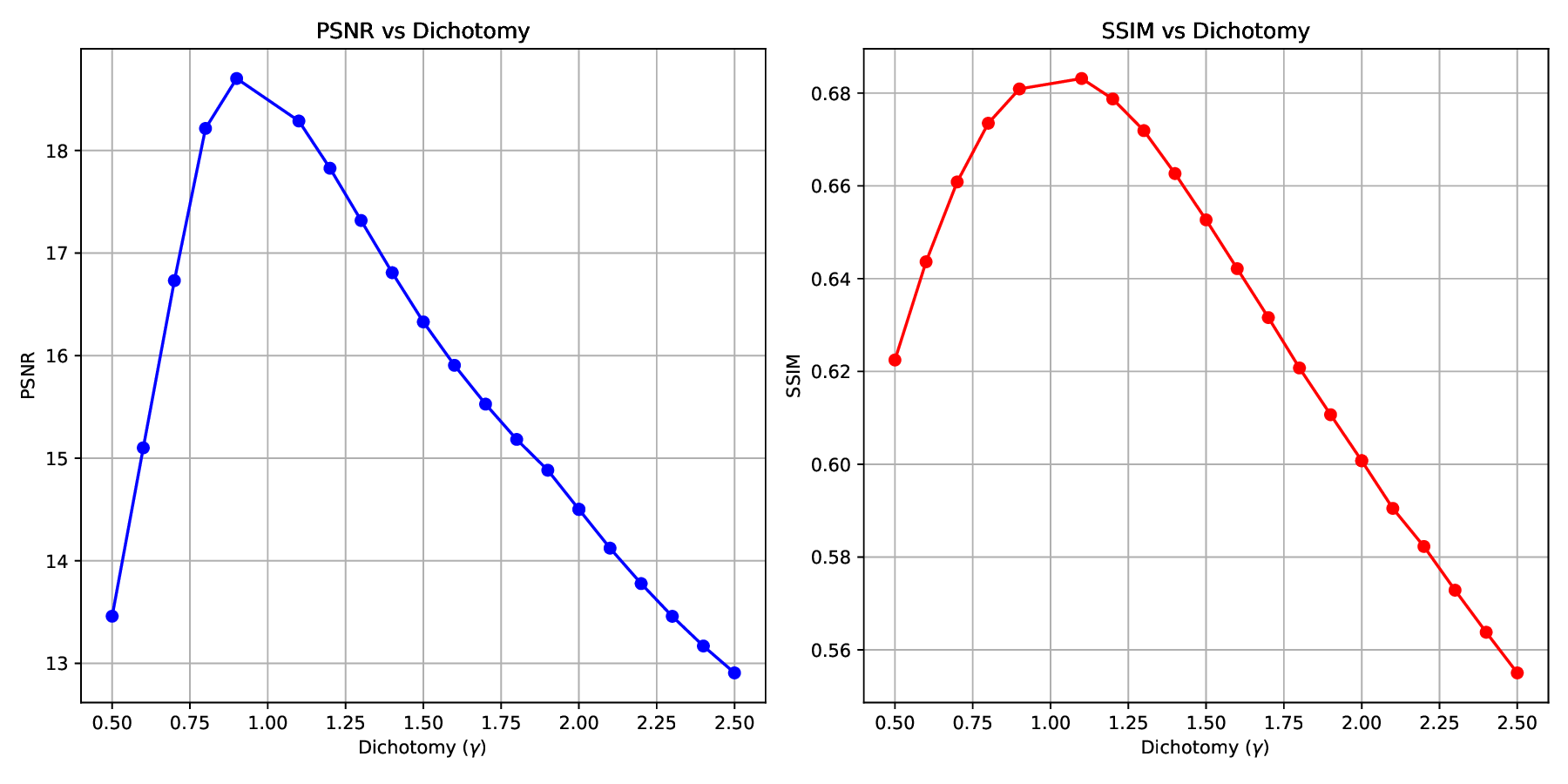}
	\caption{These plots show the curves obtained by varying $\gamma$ for the dichotomy model. Note that the $PSNR$ and $SSIM$ metrics reach their maximum values at different gamma values. The highest value reached by $PSNR = 18.7015 \pm 4.6169$ at $\gamma = 0.9$ with an associate value $SSIM = 0.6809 \pm 0.0979$, while for $SSIM = 0.6831 \pm 0.1088$, at $\gamma = 1.1$ with an associate value of $PSNR = 18.2870 \pm 4.5887$. \label{fig5}}
\end{figure}  

We compare our proposed theory with the state-of-the-art on image enhancement. The selected methods represent a rich collection of approaches ranging from conventional, convolutional neural network (CNN) learning and transformer-based methods. The methods are Adaptive Gamma Correction with Weighting Distribution (AGCWD) \cite{ref-journal5}, Natural preserved enhancement (NPE) \cite{ref-journal14}, Simultaneous reflectance and illumination estimation (SRIE) \cite{ref-journal15}, Fusion-based enhancing method (FBEM) \cite{ref-journal16}, Low light image enhancement (LIME) \cite{ref-journal17}, Deep retinex network (Retinex-Net) \cite{ref-journal18}, Learning to restore images via decomposition and enhancement(LRIDE) \cite{ref-journal19}, Semi-decoupled decomposition (SDD)
\cite{ref-journal20}, Deep local parametric filters (DeepLPF) \cite{ref-journal21}, Pre-trained Image Processing Transformer (IPT) \cite{ref-journal22}, Multi-scale residual dense network (MS-RDN) \cite{ref-journal23}, EnlightenGan (EG) \cite{ref-journal24}, Retinex-inspired unrolling with architecture search (RUAS) \cite{ref-journal25}, U-shaped transformer (Uformer)
\cite{ref-journal26}, self-calibrated illumination (SCI) \cite{ref-journal27}. These methods are highly cited and known for their robustness and the quality of their results. We evaluate our method on the widely used LOLv1 dataset \cite{ref-journal18}. LOLv1 contains 485 real low-normal pairs for training and 15 for testing. Our method did not require learning, and we used only the 15 testing images, see Figure \ref{fig4}. Authors of LOLv1 collected Low-light images by changing exposure time and ISO while fixing other camera configurations. We use the peak signal-to-noise ratio (PSNR) and the measure of structural similarity (SSIM) as quantitative metrics \cite{ref-journal28}. Table \ref{Table1} shows that our method obtains consistent best values by large margins on PSNR and second place on SSIM. Note that these numbers are obtained either by running their respective codes or from their respective papers. Figure \ref{fig5} shows the plots for different gamma values to illustrate that the maximum of both metrics is different.

\begin{table}[t]
	\scriptsize
	\centering
	\begin{tabular}{l c c c c c c c c}
	\hline
		Methods & AGCWD & NPE &  SRIE & FBEM & LIME & Retinex-Net & LRIDE & SDD\\ 
	\hline
		PSNR & 13.046 & 16.96 & 11.86 & 16.96 & 16.76 & 16.77 & 18.27 & 13.34\\
		SSIM & 0.404 & 0.481 & 0.493 & 0.505 & 0.444 & 0.56 & 0.665 & 0.635\\
	\hline	
	Methods & DeepLPF & IPT &  MS-RDN & EG & RUAS & Uformer & SCI & Dichotomy\\ 
	\hline
	PSNR & 15.28 & 16.27 & 17.20 & 17.48 & 18.23 & 16.36 & 15.80 & {\bfseries 18.7015}\\
	SSIM & 0.473 & 0.504 & 0.64 & 0.65 & {\bfseries 0.72} & 0.507 & 0.527 & {\bfseries 0.6809}\\
	\hline	
	\end{tabular}
	\caption{This table shows quantitative comparisons with state-of-the-art methods for image enhancement on the LOLv1 dataset using two standard metrics.}\label{Table1}
\end{table}

\subsection{Dichotomy Applied to the Shroud of Turin}

\begin{figure}[h]
	\centering
	\includegraphics[width=14.0cm]{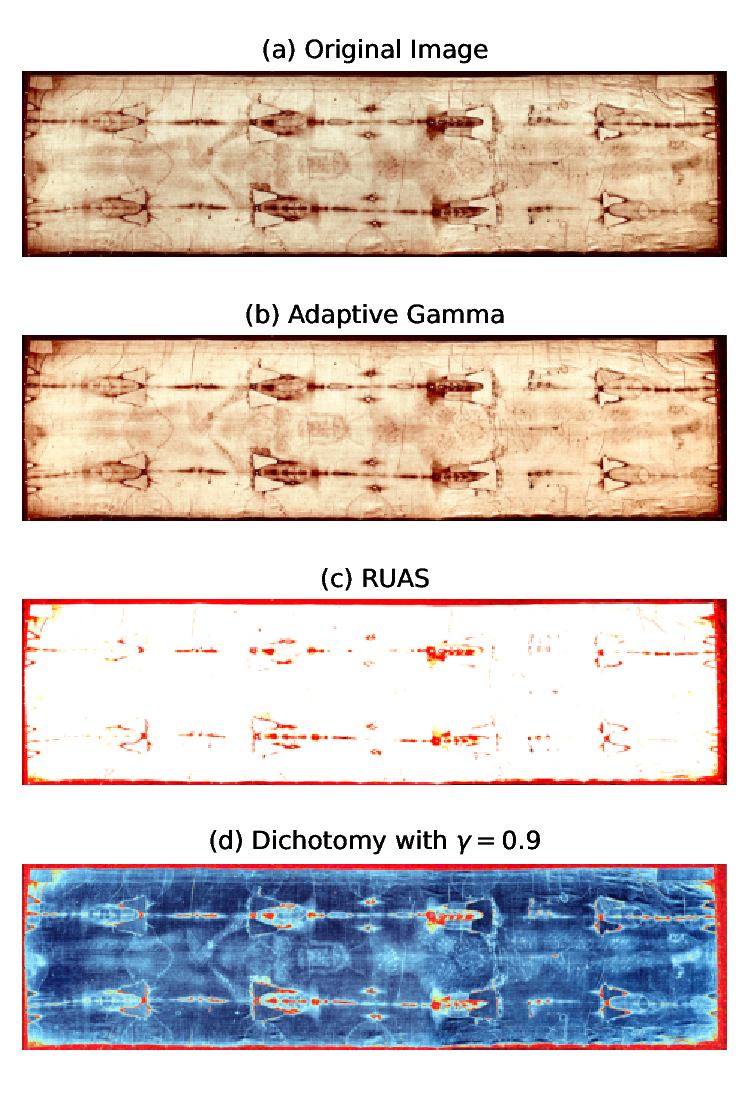}
	\caption{Unlike other image enhancement approaches, the proposed dichotomous model reveals the hidden information about the Shroud of Turin. Adaptive gamma improves the visual appearance of the original image, while RUAS erases all visual information of the man on the Shroud. \label{fig6}}
\end{figure}

The shroud of Turin is known for its negative characteristics, similar to what photographers obtain with film cameras when they need to apply chemical reactions to paper to reveal a positive image. Researchers study low-light image enhancement to improve the visual appearance of an image with poor illumination. Figure \ref{fig6} shows the original image and results after applying different algorithms. Adaptive gamma slightly modified the image since we can say there are no significant low-light problems. However, the RUAS result is catastrophic since this program erases all information. On the other hand, the proposed dichotomy model with a $\gamma = 0.9$ enhances the information hidden in the image since the model's goal is to extract the visual information. We can appreciate the whole body impress in the material while enhancing the information in comparison with a simple image inversion.

\section{Discussion}

The exposure triangle is a concept composed of three parts: aperture, shutter speed, and ISO. The exposure of an image follows strict rules that photographers and anyone interested in image formation, like computer vision, photogrammetry, and image processing, must understand and follow closely to ensure consistency across shots. Mastering the exposure triangle means balancing all three elements since adjusting one forces you to compensate through the other two to achieve an acceptable exposure. The model presented in this article allows us to update an image that is considered unuseful or to extract information in real time despite the lack of light. There are numerous situations in which the model can help change the rules of photography. One positive aspect is the return to mathematical modeling since such analysis follows deductive reasoning and, therefore, is subject to rigorous explanation, which contrasts with modern approaches like deep neural networks rooted in inductive reasoning with data dependency problems.

Today, scientists and engineers use neural networks to design powerful heuristics for complex pattern recognition problems, although not all of us can run large models or train them. The goal of doing more with less should be a focus of artificial intelligence. As in the early years, we have placed hope in symbolic models, and as we have seen in this work, simple rules prove effective for problems of a specific kind, such as tone in images. The proposal presented in this paper follows mathematical modeling through a rigorous and methodological approach that researchers can combine with symbolic and subsymbolic learning methods. The symbolic representation presented in our work is less demanding and helps create powerful tools that are more widely available. This work represents an effort to confront the overwhelming idea of deep learning by advocating for a methodology to develop understandable and justifiable explanations so that not all computer science starts from such technologies. 
The results reported on an image enhancement problem show that the model obtains practical results against generative and transformer models. A well-taken photograph requires specific parameters and not a model containing all possible parameters for all possible images. Image enhancement requires a comprehensive study that formulates the problem under the idea presented here, and we will develop this in future research.

\section{Conclusions and Future Work}

The importance of the theorem \ref{theorem} is presented in the conservation of information since the operations are punctual and the neighborhood does not intervene in the calculations. The proposed operation does not modify aspects related to geometry and morphology linked to the location, orientation, shape, and number of characteristics. Therefore, the equation does not change the representation of information (brightness and contrast), there being a direct relationship between $f_{in}$ and $f_{out}$, since the transformation is invertible; see Lemmas \ref{Lemma8} and \ref{Lemma14}. This property allows the extraction of reliable information based on the creation of a space for the representation of $f_{out}$ with different values of $\gamma$, which allows the contrast of other tones locally by highlighting edges, points, lines, and curves helping to identify essential elements for higher level computer vision tasks.

Note that Lemmas \ref{Lemma6}, \ref{Lemma7}, and \ref{Lemma12} allow us to prove that for a given input range of values $f_{in}$, the output $f_{out}$ will have the same range for any value of $\gamma$, so it is clear that it is a stable model, fast in terms of calculation operations and easy to implement. Since the calculations are at the pixel level, one can use look-up tables (LUTS) when the input pixel values are integers or use $f_{out}$ directly when the pixels are real numbers, see Lemma \ref{Lemma14}. In practice, to determine the gamma value, we can identify three cases with the help of the histogram:
1) when the image is underexposed, the positive slope of $f_{out}$ has a high magnitude when gamma tends to zero and is less than 0.9 because the pixels within the reference image are approaching pure black; see Lemma \ref{Lemma9} and Lemma \ref{Lemma14}; 2) When the image is overexposed the negative slope of $f_{out}$ has a high magnitude when gamma is higher than two because the pixels within the reference image are approaching pure white; see Lemma \ref{Lemma9} and Lemma \ref{Lemma14}; 3) in the mixed case, where both cases appear, the gamma values considered above are not necessarily the same but can overlap, and a more comprehensive gamma range between $0.8$ and $4$ is more suitable. The transformation allows increasing the values of the underexposed regions in their positive slope and decreasing the values of the overexposed areas of the negative slopes; see Lemma \ref{Lemma9}, Lemma \ref{Lemma11} and Lemma \ref{Lemma14}. Note that each image can be considered an instance of the case studied. Hence, the values proposed here serve only as a reference and would need to be adjusted with the image histogram to propose a gamma value that is more suitable for the final task.

This article introduces a new mathematical modeling of tone to recover the contrast of image information. It opens new opportunities to study images from the viewpoint of invariant theory.

\begin{Definition}{\bfseries[Invariant principle]} An invariant is a numerical quantity calculated from a certain configuration. It has the property of remaining unchanged even if the initial configuration undergoes a certain kind of transformation.
	To be more precise, consider two sets $E$ and $F$, and a set of transformations of $E$ into $F$. $I$ being a function of elements of $F$. $I$ is an invariant for $T$ if it takes the same value for all images of an element of $E$:
	
	\begin{equation}
		\forall e \in E,\;\; \forall t,\;\; t' \in T \;\;\; I(t(e)) = I(t'(e))
	\end{equation}
	\noindent
	where e is an element of the set E and t, t' are transformations of E into F.
\end{Definition}

The power law is a fundamental equation for numerous image analysis and processing tasks, such as image enhancement, that deserves utterly new work. We leave future work to develop theoretical and practical feature extraction and description models. We expect that the mathematical model introduced in this article will be of help for tasks such as perceptual uniformity, detail contrast, edge detection, interest point detection, accurate image analysis, lightning information, visual attention, image classification, and speeding processing power, which are helpful in computational photography and image processing but also in computer vision, robotics, and artificial intelligence in general.

\section{Patents}
		
Gustavo Olague and Axel Mart\'{\i}nez. Sistema para el tratamiento de im\'{a}genes sin información visible al ojo humano. Instituto Mexicano de la Propiedad Intelectual. Solicitud de patente MX/a/2024/004585.
		
		\vspace{6pt} 
		
		
		
		

\section{Acknowledgments}{The following project support this research: Project titled "MOVIE: Modelado de la visi\'{o}n y la evoluci\'{o}n" CICESE-634135.}

\end{document}